\documentclass[conference]{IEEEtran}
\IEEEoverridecommandlockouts
\usepackage{hyperref}
\usepackage{cite}
\usepackage{amsmath,amssymb,amsfonts}
\usepackage{algorithmic}
\usepackage{graphicx}

\usepackage{textcomp}
\usepackage{xcolor}
\usepackage{url}
\usepackage{subfigure}
\def\BibTeX{{\rm B\kern-.05em{\sc i\kern-.025em b}\kern-.08em
    T\kern-.1667em\lower.7ex\hbox{E}\kern-.125emX}}
\usepackage{tikz}
\usetikzlibrary{shapes,arrows,fit}
\usetikzlibrary{positioning}
\usetikzlibrary{decorations.pathreplacing}
\usetikzlibrary{calc}
\PassOptionsToPackage{final, activate={true, nocompatibility}, tracking=true, kerning=true, factor=1100, stretch=10, shrink=10, babel=true}{microtype}

\usepackage[belowskip=-15pt,aboveskip=0pt, font=small]{caption}
\usepackage{microtype}
\usepackage{pgfplots}
\usepackage{csvsimple}

\usepackage{pgfplotstable,filecontents}
\pgfplotsset{compat=1.9}
\usepackage{mathrsfs}
\usepackage{bbm}
\usepackage{float}
\usepackage{mathtools,}
\usepackage{amsthm}

\usepackage{multicol}
\usepackage{xurl}
\usepackage{amsmath}
\usepackage{mathrsfs}
\usepackage[scleft,ruled,linesnumbered]{algorithm2e}

\newtheorem{property}{\begin{small}Property\end{small}}

\pgfplotsset{compat=1.16}
\newcommand{\graphScale}{0.7}

\newtheorem{theorem}{\begin{small}Theorem\end{small}}

\newcommand{\pluseq}{\mathrel{+}=}
\DeclareMathOperator*{\argmax}{arg\,max}

\begin{document}

\title{Budgeted Classification with Rejection: An Evolutionary Method with Multiple Objectives}
\author{\IEEEauthorblockN{Nolan H. Hamilton and Errin W. Fulp} \IEEEauthorblockA{Department of Computer Science\\
Wake Forest University, Winston-Salem, NC, USA \\
Email: haminh16@wfu.edu, fulp@wfu.edu}}
\renewcommand{\IEEEbibitemsep}{0pt plus 1pt}
\makeatletter
\IEEEtriggercmd{\reset@font\normalfont\footnotesize}
\makeatother
\IEEEtriggeratref{1}
\maketitle
\captionsetup[table]{skip=2pt}
\begin{abstract}
Classification systems are often deployed in resource-constrained settings where labels must be assigned to inputs on a budget of time, memory, etc. Budgeted, sequential classifiers (BSCs) address these scenarios by processing inputs through a sequence of partial feature acquisition and evaluation steps with early-exit options. This allows for an efficient evaluation of inputs that prevents unneeded feature acquisition. To approximate an intractable combinatorial problem, current approaches to budgeted classification rely on well-behaved loss functions that account for two primary objectives (processing cost and error). These approaches offer improved efficiency over traditional classifiers but are limited by analytic constraints in formulation and do not manage additional performance objectives. Notably, such methods do not explicitly account for an important aspect of real-time detection systems---the fraction of ``accepted'' predictions satisfying a confidence criterion imposed by a risk-averse monitor.

We propose a problem-specific genetic algorithm to build budgeted, sequential classifiers with confidence-based reject options. Three objectives---accuracy, processing time/cost, and coverage---are considered. The algorithm emphasizes Pareto efficiency while accounting for a notion of aggregate performance via a unique scalarization. Experiments show our method can quickly find globally Pareto optimal solutions in very large search spaces and is competitive with existing approaches while offering advantages for selective, budgeted deployment scenarios. 
\end{abstract}
\section{Introduction}
    Many real-world classification scenarios present latency constraints between observation of inputs and label assignment, and a common
    difficulty regards balancing processing cost with accuracy \cite{ji2007,trapeznikov2013,xu2012}.
    In these \textit{budgeted} learning settings, the magnitude of resources (e.g., time, memory) expended while evaluating inputs represents the \textit{cost of classification}, where a greater cost of classification is generally associated with improved classifier performance.

    Cost of classification can be specified further as the sum of \textit{feature acquisition cost} and \textit{classifier evaluation cost}. Feature acquisition cost addresses the resources spent generating features for classification that are not present during the initial observation of inputs at test-time. With features acquired, classifier evaluation cost measures the resources exhausted while assigning labels to inputs using a trained classifier and is often negligible compared to feature acquisition cost \cite{kusner}.
    
    As an example scenario warranting a budgeted approach, consider the task of classifying e-mail messages as spam. For timeliness and users' security, a large influx of messages must be evaluated quickly, and informative features such as subject-line character diversity, sender reputation score, sender location, URL count, etc. can require considerable cpu-time to compute. Generating all of these features for every input may therefore impose unacceptable processing cost. Fortunately, it is often possible to use small, cheap subsets of features to correctly classify a considerable fraction of test-time inputs ~\cite{trapeznikov2013,xu2012,xu2014,ji2007,nan16}, and \textit{budgeted, sequential classifiers} leverage this property to reserve intensive processing for only the inputs that require it. Using this design, a set of classifiers with varying feature sets are arranged in a  sequence of stages to evaluate inputs~\cite{trapeznikov2013,ji2007,wang2014}. Once an input receives a confident class prediction in a particular stage, processing ceases and no cost is incurred for the remaining stages' features that were not used. Specific use-cases of sequential, budgeted learning can be found across a diverse range of application domains \cite{hamilton20202,nogueira2019}. However, while budgeted learning schemes present an opportunity for efficient processing of inputs in time-critical instances, there are a multitude of important design considerations that can affect performance of these protocols, making configuration a difficult optimization problem that has received significant attention in the past decade \cite{trapeznikov2013, ji2007, xu2014, xu2012, nan16,jansich}. 
    \section{Related Work}
    A variety of budgeted approaches have been proposed in the past decade. In this section, we conduct a brief survey of some prominent algorithms. In general, budgeted classifiers balance the cost-accuracy trade-off by minimizing a well-behaved objective function that is increasing with respect to error and cost of classification. Note, there are several loosely-related learning paradigms that will not be addressed in this manuscript. \textit{Classifier cascades}, for example, leverage a fundamental assumption of class imbalance and do not assign positive labels at intermediate stages \cite{viola}. 
    
    The authors in \cite{trapeznikov2013} formulate the problem with \textit{early-exit} options at each stage that prevent unnecessary feature acquisition by ceasing processing when early-stage predictions are deemed conclusive. To avoid difficult combinatorial aspects, the order of features is fixed beforehand. The authors construct and minimize a global, smooth cost function by coordinate descent and can thus guarantee local optimality of solutions---but only for the chosen ordering of features and stages. Wang et al. \cite{wang2014} offers improved theoretical guarantees relative to the approach proposed in \cite{trapeznikov2013} by formulating the problem in a convex framework (linear program). The authors are then able to guarantee globally optimal solutions under their approximate formulation and a fixed feature/stage order. 
 
    Several existing methods incorporate feature selection/order implicitly. \cite{xu2012} proposes \textit{GreedyMiser}---a feature-budgeted variant of stage-wise regression \cite{friedman2001}. Limited-depth regression trees are used as weak learners and are constructed with a modified impurity function accounting for cost of feature extraction. These weak learners are combined to form a final classifier. For its simplicity and efficacy, GreedyMiser has become one of the most popular and best-cited feature-budgeted approaches to classification and is frequently used as a benchmark \cite{xu2014,nan16,nan2017,andrade20}. \cite{xu2014} proposes \textit{Cost-Sensitive Tree of Classifiers}; This method builds a budgeted tree of classifiers with leaf nodes optimized for a specific subset of the input space.

    While the described methods offer a marked improvement in efficiency over non-budgeted classification of inputs in resource-constrained settings, several limitations are consistent throughout this body of work. First, many of the methods do not optimize the order of the features/stages \cite{trapeznikov2013,wang2014} and instead resort to ``increasing cost'' heuristics to choose an ordering of features (expensive features placed at later stages). But this can be inefficient if, for instance, a large fraction of cheap features are uninformative. In addition, intuitively modeling complex feature interactions beforehand can prove difficult, and such interactions can prove consequential if variables are more/less discriminative when grouped together \cite{ji2007}. 
    
    Another common theme of the methods discussed above is that they exploit a cost/accuracy trade-off in which cheaper solutions are generally less accurate than costly solutions. Unfortunately, the decreased accuracy of the cheaper classifiers can restrict applicability in critical, real-world environments. One possible remedy to increase accuracy of classifiers is to apply a \textit{reject option}
    that rejects low-confidence predictions (i.e., inputs with unconfident predictions are discarded). These \textit{selective classifiers} have been studied independently from budgeted classification and seek to balance the \textit{coverage}-accuracy tradeoff, where coverage refers to the expected fraction of inputs that are \textit{not} rejected \cite{yaniv10,geifman2017}. Existing budgeted approaches do not apply reject options, but perfect coverage is often unnecessary in practical scenarios \cite{geifman2017}, and conservative decision makers with cost constraints may benefit in sacrificing coverage (rather than processing efficiency) for improved accuracy.

    \section{\texttt{EMSCO}---\textit{Evolutionary Multi-Stage Classifier Optimizer}}
    The limitations discussed in the previous section motivate a budgeted \textit{and} selective protocol accounting for accuracy, cost, and coverage during optimization. We propose such a method with the aim of:
    \begin{enumerate}
        \item Offering greater accuracy than existing budgeted methods by rejecting uncertain predictions
        \item Matching or reducing processing cost compared to existing budgeted methods
        \item Preserving high coverage
    \end{enumerate}

    A problem-specific genetic algorithm serves as the fundamental mechanism for optimization. Such genetic algorithms (GAs) employ a population-based approach to optimization in which the generation/population $G_1$ is a product of mutation, selection, and crossover on the solutions in $G_0$. GAs are noted for their ability to find optimal or near-optimal solutions for very large combinatorial problems in polynomial time and properly manage multiple objectives in a Pareto efficient manner \cite{deb01}. These algorithms also grant significant mathematical flexibility in formulation, as there is no need for objectives to be smooth or otherwise well-behaved. 
     In our setup, three objectives are considered: coverage ($g_1$), accuracy ($g_2$), and (inverse) feature acquisition cost ($g_3$)  (See Section \ref{objectives} for a detailed description of these objectives.) These objectives are optimized over a set of feasible stage designs, defined in Section \ref{sec:solspace} as ordered partitions of the feature set. 
     
     Note, we move several supplementary resources (pseudo- code, demos, additional experiments, etc.) that readers may find informative to a public repository\footnote{Available at \url{https://github.com/nolan-h-hamilton/EMSCO-supplement}}.

\subsection{Problem Setting} \label{sec:setup}
    As in \cite{trapeznikov2013}, we assume a set of training examples from past instances for which measurements of all features $\mathcal{F}_1, \mathcal{F}_2, \ldots, \mathcal{F}_n$ and correct labels are available. The aim is to find system designs that reduce feature acquisition cost and maintain high accuracy and coverage at test/prediction time.
    
    Upon initial observation at test-time, input $\mathbf{x}$ begins with no acquired features, and features are thereafter attained as needed through a sequence of stages. That is, stage $j$ acquires a subset of features $Q_j \subseteq \mathcal{F}$ to evaluate\footnote{For its generalization ability, computational efficiency, and accurate class probability estimates, \texttt{EMSCO} employs $L_2$-regularized logistic regression ($\lambda=1$) at each stage in this paper. \texttt{EMSCO}'s performance can be improved by using more complex classification methods (e.g., Random Forest) at each stage, but this may result in markedly increased training time for large datasets.} input $\mathbf{x}$.  With these features acquired, the stage-$j$ classifier, $\mathscr{C}(Q_j,\mathbf{x})$, learned beforehand on features in $Q_j$ during the training phase, returns a set of prediction confidences $\mathscr{P_{\mathbf{x_j}}} = \{p_1, p_2, \ldots, p_l\}$ corresponding to possible labels for input $\mathbf{x}$. Note, every feature has a corresponding cost of acquisition given by the set $\mathcal{C}$---acquiring feature $\mathcal{F}_i$ for evaluation of $\mathbf{x}$ increases the cost of classification incurred by the input by $\mathcal{C}_i$ units.
    
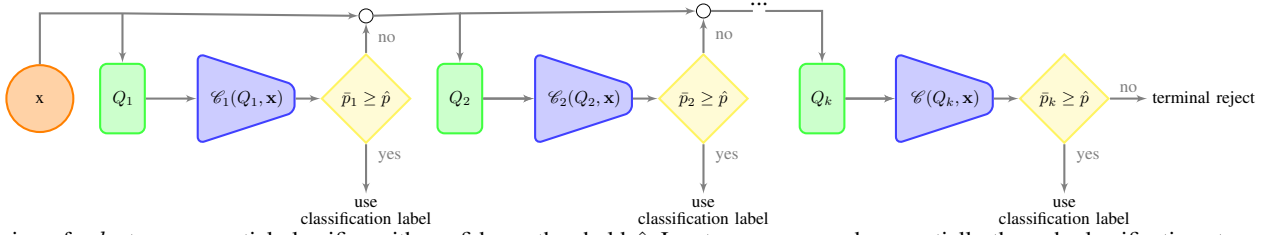
\begin{figure*}[t] 
\begin{center}
\resizebox{\textwidth}{!}
{ 
\tikzstyle{decision} = [diamond, very thick, draw=yellow!75, fill=yellow!20,
    text width=4.25em, text badly centered, node distance=3.2cm, inner sep=0pt]
\tikzstyle{block} = [rectangle, very thick, draw=green!75, fill=green!20,
    text width=6em, text centered, rounded corners, minimum height=4em]
\tikzstyle{stage} = [rectangle, very thick, draw=black, fill=blue!20,
    text width=5em, text centered, rounded corners, minimum height=4em]
\tikzstyle{feature} = [rectangle, very thick, draw=green!75, fill=green!20,
    text width=2em, text centered, rounded corners, minimum height=4em]
\tikzstyle{trap} = [trapezium, draw, minimum width = 1.5cm,
trapezium left angle = 108, trapezium right angle = 108, shape border rotate = 90, very thick, draw=black!75, fill=black!20, text width=5em, text centered, rounded corners, minimum height=4em]
\tikzstyle{data} = [trapezium, draw, trapezium left angle=60,
  trapezium right angle=-60, very thick, draw=green!75, fill=green!20,
    text width=5em, text centered, rounded corners, minimum height=4em]
\tikzstyle{line} = [draw, very thick, color=black!50, -latex']
\tikzstyle{cloud} = [draw=black!75, circle, very thick, fill=black!20, node distance=2.5cm,
    minimum height=3.85em]
\tikzstyle{decision} = [diamond, very thick, draw=yellow!75, fill=yellow!20,
    text width=4.25em, text badly centered, node distance=3.2cm, inner sep=0pt]
\tikzstyle{block} = [rectangle, very thick, draw=green!75, fill=green!20,
    text width=6em, text centered, rounded corners, minimum height=4em]
\tikzstyle{stage} = [rectangle, very thick, draw=black, fill=blue!20,
    text width=5em, text centered, rounded corners, minimum height=4em]
\tikzstyle{feature} = [rectangle, very thick, draw=green!75, fill=green!20,
    text width=2em, text centered, rounded corners, minimum height=4em]
\tikzstyle{trap} = [trapezium, draw, minimum width = 1.5cm,
trapezium left angle = 108, trapezium right angle = 108, shape border rotate = 90, very thick, draw=blue!75, fill=blue!20, text width=5em, text centered, rounded corners, minimum height=4em]
\tikzstyle{data} = [trapezium, draw, trapezium left angle=60,
  trapezium right angle=-60, very thick, draw=green!75, fill=green!20,
    text width=5em, text centered, rounded corners, minimum height=4em]
\tikzstyle{line} = [draw, very thick, color=black!50, -latex']
\tikzstyle{cloud} = [draw=orange!100, circle, very thick, fill=orange!35, node distance=2.5cm,
    minimum height=3.85em]

\begin{tikzpicture}[node distance = 2.5cm and 1.0cm, auto, inner sep = 1mm, scale=.75]

   \node [cloud] (start) {\shortstack{x}};
   \node [feature, right = 0.5 of start] (F1) {{$\Large{Q_1}$}};
   \coordinate[above = 1cm of F1] (d1);
   \node [trap, right of = F1] (stage1){$\mathscr{C}_1(Q_1,\mathbf{x})$};
   \node [decision, right = 0.5cm of stage1] (decision1) {$\bar{p}_{1} \geq \hat{p}$};  
   \node [draw, circle, above = 0.6cm of decision1] (continue1) {};
   \node [below = 1cm of decision1] (done1) {\shortstack{use \\ classification label}};

   \node [feature, right = 0.5cm of decision1] (F2) {{$\Large{Q_2}$}};
   \coordinate[above = 1.0cm of F2] (d2);
   \node [trap, right of = F2] (stage2) {$\mathscr{C}_2(Q_2,\mathbf{x})$};
   \node [decision, right = 0.5cm of stage2] (decision2) {$\bar{p}_{2} \geq \hat{p}$};  
   \node [draw, circle, above = 0.7cm of decision2] (continue2) {};
   \node [below = 1cm of decision2] (done2) {\shortstack{use \\ classification label}};

   \coordinate[right = 0.85cm of continue2] (upper_dot_start);
   \coordinate[right = 0.25cm of upper_dot_start] (upper_dot_end);
   \coordinate[right = 0.25cm of decision2] (dot_start);
   \coordinate[right = 0.25cm of dot_start] (dot_end);

   \node [feature, right = 0.5 of dot_end] (Fn) {{$\Large{Q_k}$}};
   \coordinate[above = 1.05cm of Fn] (dn);
   \node [trap, right = 1.0cm of Fn] (stagen) {$\mathscr{C}(Q_k,\mathbf{x})$};
   \node [decision, right = 0.5cm of stagen] (decisionn) {$\bar{p}_{k} \geq \hat{p}$};  
   \node [below = 1cm of decisionn] (donen) {\shortstack{use \\ classification label}};
   \node[right = 0.75cm of decisionn] (idk) {\shortstack{terminal reject}};

   \path [line] (start) |- (d1) -- (continue1);
   \path [line] (d1) -- (F1);
   \path [line] (F1) -- (stage1);
   \path [line] (stage1) -- (decision1);
   \path [line] (decision1) -- node[swap, midway] {~no} (continue1);

   \path [line] (F2) -- (stage2);
   \path [line] (continue1) -- (d2) -- (F2);
   \path [line] (d2) -- (continue2);
   \path [line] (decision1) -- node[near start] {~yes} (done1);
   \path [line] (stage2) -- (decision2);
   \path [line] (F2) -- (stage2);
   \path [line] (decision2) -- node[swap, midway] {~no} (continue2);
   \path [line] (decision2) -- node[near start] {~yes} (done2);

   \path (upper_dot_start) -- node[]{\Large{...}} (upper_dot_end);
   \path (dot_start) -- (dot_end);

   \path [draw, very thick, color=black!50] (continue2) -- (upper_dot_start);

   \path [line] (Fn) -- (stagen);
   \path [line] (Fn) -- (stagen);
   \path [line] (upper_dot_end) -- (dn) -- (Fn);
   \path [line] (stagen) -- (decisionn);
   \path [line] (decisionn) -- node[midway] {no} (idk);
   \path [line] (decisionn) -- node[near start] {~yes} (donen);

\end{tikzpicture}

}
\caption{Design of a $k$-stage sequential classifier with confidence threshold $\hat{p}$. Inputs are processed sequentially through classification stages until they achieve a class probability greater than $\hat{p}$ or all features have been exhausted. \label{fig:workflow}}

\end{center}

\end{figure*}

    We assume a \textit{risk-averse monitor} has decided to accept only confident predictions to improve accuracy. At preterminal stages $j < k$, an early-exit decision is made to determine whether the input prediction is sufficiently confident to be accepted or if additional processing is necessary. Let $\bar{p_j} =  \max \mathscr{P_{\mathbf{x_j}}}$ and $\hat{p}$ be a confidence threshold specified by the monitor. At stage $j<k$, if $\bar{p_j} < \hat{p}$, the input is sent to stage $j+1$ where it is evaluated with feature set $Q_{j+1}$ (Note, $Q_j \subset Q_{j+1}$). Conversely, if $\bar{p_j} \geq \hat{p}$, processing ceases, and $\mathbf{x}$ is assigned the label corresponding to the maximum class probability.
     At the final stage $k$, if $\bar{p_k} < \hat{p}$, evaluation of $\mathbf{x}$ is deemed \textit{inconclusive} with no label assigned. In this case, we say that $\mathbf{x}$ has been \textit{rejected}. This is a safe but generally undesirable result as it decreases utility of the system---while wanting high accuracy, the above-mentioned risk-averse monitor also desires that the system yield insight. With a confidence threshold specified \textit{a priori} according to a system monitor's preferences and misclassification penalties, we aim to find a stage configuration that provides high accuracy, low processing cost, and high coverage. Figure \ref{fig:workflow} offers a visual depiction of sequential classification with early-exits and a reject option.

    \subsection{Solution Space}
    \label{sec:solspace}
    Let $k \in \mathbb{N}$ with $k < n$. Solutions are $k$-\textit{partitions} of feature set $\mathcal{F}$. That is, $\bigcup_{j=1}^{k} {Q_j} = \mathcal{F}$ where $Q_j \neq \emptyset$ denotes the features acquired in stage $j$. Due to the sequential nature of BSCs, the ordering of the stages must also be considered, and the number of ordered $k$-partitions $|\mathcal{P}_{(n,k)}|$ on a feature set with dimension $n$ is computed by multiplying the number of unordered partitions by $k!$:
    \begin{equation} 
     {|\mathcal{P}_{(n,k)}|} = k! \cdot S_2(n,k) = \sum\limits_{i=0}^{k}(-1)^{k-i}{k \choose i}i^n,
    \end{equation}
    where $S_2(n,k)$ represents the number of unordered $k$-partitions of a set with $n$ elements, or a \textit{Stirling number of the second kind}.
    
    To consider a range of possible stage counts, we define the solution space $\mathcal{S}_{(n,k)}$ as the set of all solutions with \textit{up to and including} $k$ stages. That is, 
    \begin{equation} \label{solutionspace}
    \mathcal{S}_{(n,k)} = \bigcup_{j=1}^{k}  \mathcal{P}_{(n,j)}.
    \end{equation}
    For instance, $\mathcal{S}_{(n,k=3)}$ includes solutions with one, two, or three stages.
    \begin{theorem} \label{thm:solutionspace}
      (\textit{Size of Search-Space}) For fixed $k \in \mathbb{N}$, the number of feasible solutions $|\mathcal{S}_{(n,k)}|$ is asymptotically equivalent to $k^n$.
     \begin{proof}
     Since there is no intersection between terms on the RHS of (\ref{solutionspace}), we have $\sum_{j=1}^{k} |\mathcal{P}_{(n,j)}| = |\mathcal{S}_{(n,k)}|.$
     Taking the limit of ratio $\frac{|\mathcal{P}_{(n,j)}|}{j^n}$ as $n \rightarrow \infty$ yields $|\mathcal{P}_{(n,j)}| \sim j^n$. 
     We can then write:
    \begin{gather*}
     \lim_{n\to\infty} \frac{1}{k^n} |\mathcal{S}_{(n,k)}| =  \lim_{n\to\infty} \frac{1}{k^n}\sum_{j=1}^{k}  |\mathcal{P}_{(n,j)}| 
     =\lim_{n\to\infty} \frac{1}{k^n}\sum_{j=1}^{k} j^n\\ = \lim_{n\to\infty} \left(\frac{k}{k}\right)^n = 1.~ \textup{That is,}~ |\mathcal{S}_{(n,k)}| \sim k^n.
    \end{gather*}
    \end{proof}
    \end{theorem}
In practice, all features in stage $j$ are automatically appended to stage $j+1$ as they are already acquired and can be referenced for no cost. However, this technicality does not affect the above analysis, since it depends only on the unique elements acquired at each stage.

 As our defining structure for BSCs, ordered feature set partitions possess several beneficial properties. In many cases, certain features are uninformative alone but become highly discriminative when evaluated together \cite{ji2007}. By allowing stages to contain multiple features, these scenarios are acknowledged implicitly during optimization. Additionally, using the representation proposed in Section \ref{representation}, ordered feature set partitions can be conveniently modeled as chromosomes.
 
    \subsection{Objectives}\label{objectives}
\begin{figure}

  \includegraphics[scale=0.3]{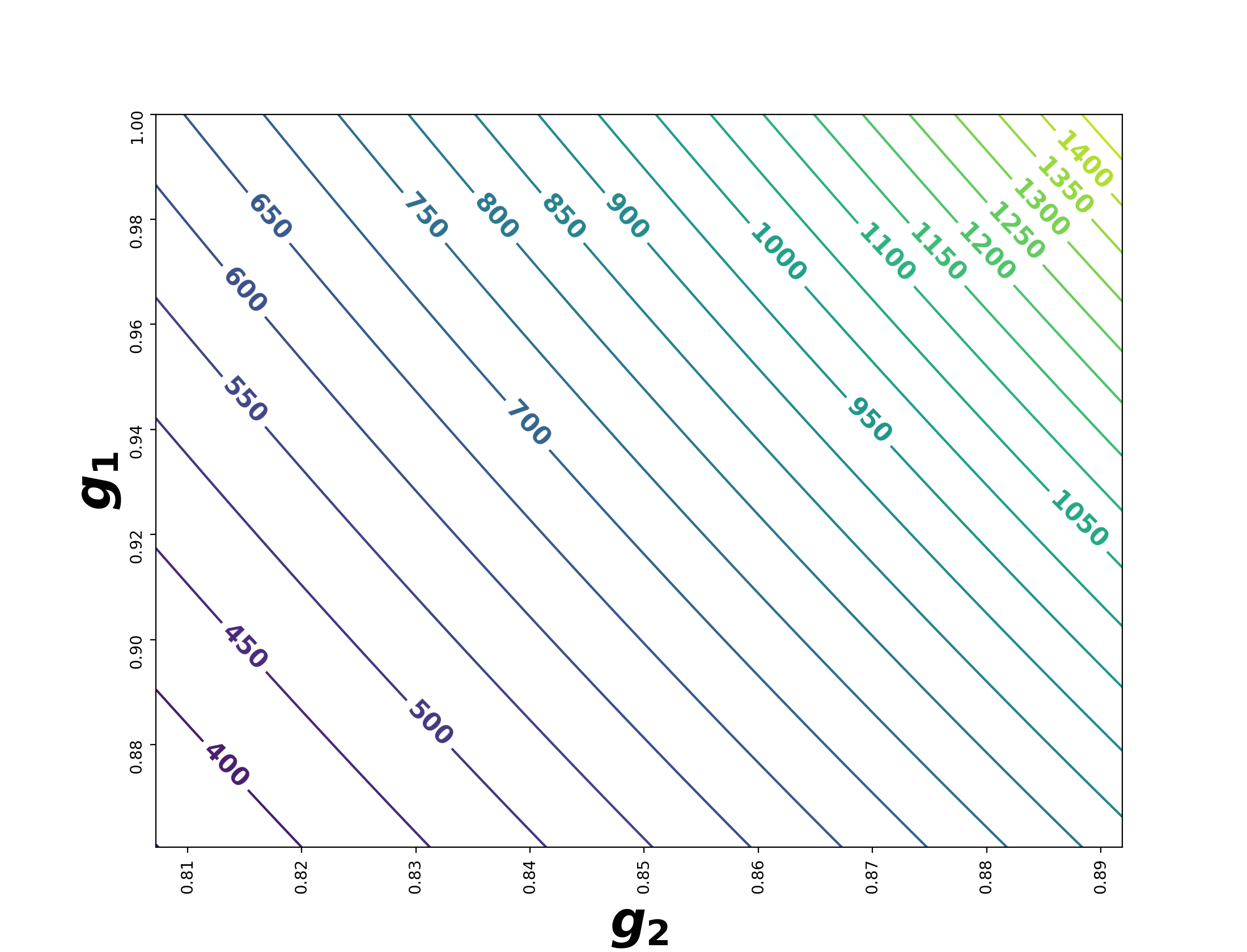}

\begin{center}
\caption{\begin{footnotesize}\textit{Performance Trade-offs}. A smooth approximation of the fitness landscape for the Credit data set (See Section \ref{datasets}) with $k=4$ exhibiting the trade-offs between cost, coverage, and accuracy. Color is determined by cost ($g^*_3$), with the yellow regions corresponding to solutions with greater cost. Cost is greatest where accuracy and coverage are highest. \end{footnotesize}}
  \label{fig:credit_landscape}
\end{center}

\end{figure}

    Objectives are designed to incorporate important performance aspects of a real-time classification system.  The relationship between objectives is explored visually in Figure \ref{fig:credit_landscape}. Note, in this subsection, $m$ denotes the index of the final stage at which input $\mathbf{x}$ is processed. $\hat{\mathcal{X}}$ is a held-out subset of data used to estimate performance of solutions.
    \subsubsection{Coverage} \label{sec:conc}
     In our setup, coverage is measured as the fraction of predictions that meet a targeted confidence threshold, $\hat{p}$.
     Several methods have been proposed to determine reject decisions \cite{geifman2017}, but we utilize confidence-based reject decisions because they are frequently used in practice \cite{cancerconf,patel} and efficient to compute.
    \begin{equation}\label{conc}
    g_1(Q) = \frac{1}{N}\sum_{\mathbf{x} \in \hat{\mathcal{X}}} \mathbbm{1}[\bar{p_m} \geq \hat{p}]
    \end{equation}
    If coverage were not considered during optimization, a risk-averse monitor accepting only confident predictions may obtain little information from the resulting configuration since a system rejecting the vast majority of inputs (but correctly assigning labels to a few) could be favored. We assume system monitors desire accuracy but also wish that the classification system yield sufficient insight and does not reject an excessive fraction of predictions.
    \subsubsection{Accuracy}
     As in selective classification \cite{yaniv10}, we measure accuracy as the proportion of \textit{accepted and correctly classified} inputs with respect to the total number of accepted predictions. Let $y$ denote the true label for $\mathbf{x} \in \hat{\mathcal{X}}$, and $\bar{\mathscr{C}_m}(Q_m,\mathbf{x})$ return the label corresponding to the maximum class probability after evaluation in stage $m$.  For $N^* = N\cdot g_1(Q)$, accuracy is measured as:
    \begin{equation} \label{acc}
    g_2(Q) =   \begin{cases} 
      {\frac{1}{N^*} \sum\limits_{\scriptsize{\mathbf{x} \in \hat{\mathcal{X}}}} \mathbbm{1}[\bar{\mathscr{C}_m}(Q_m,\mathbf{x}) = y \And \bar{p_m} \geq \hat{p}]} & N^* > 0, \\
    0 &  N^* = 0
   \end{cases}
    \end{equation}
    The case $N^* = 0$ was not realized in this paper's experiments.
    \subsubsection{Cost} \label{sec:cost}
    Because feature acquisition typically comprises the bulk of test-time processing expense \cite{kusner}, we disregard classifier evaluation cost and measure cost $g_3^*(\cdot)$ as the average sum of acquisition costs per test input:
    $$g_3^*(Q) = \frac{1}{N}\sum_{\mathbf{x} \in \hat{\mathcal{X}}} \left(\sum_{j = 1}^{m} \mathcal{C}_{Q_j}\right),$$
    where $\mathcal{C}_{Q_j}$ denotes the sum of feature costs acquired in stage $j$. Note that the cost of rejected records is included since the system requires resources to arrive at an inconclusive result.
    
    To maintain consistency as a maximization problem and a $(0,1]$ or $[0,1]$ scale for objectives, the minimum raw cost of all solutions in the current population is determined and used for normalization. Then, for $Q \in G_h$, \textit{inverse cost} $g_3(Q)$, is measured as:
    \begin{equation}\label{cost} 
        g_3(Q) = \frac{\min_{U \in G_h} g^*_3(U)}{g^*_3(Q)}
    \end{equation}
   
    \subsection{Problem Definition} \label{problemdef}
     \texttt{EMSCO} addresses the following global multi-objective optimization by seeking solutions that are \textit{non-dominated} \cite{deb01} in the global solution space $\mathcal{S}_{(n,k)}$:
    \begin{equation} \label{problem}
        \argmax_{Q \in \mathcal{S}_{(n,k)}} \left(g_1(Q),g_2(Q),g_3(Q)\right).
    \end{equation}
    In this manuscript, solutions to this problem are referred to as ``globally optimal'' or ``globally non-dominated''. 
    \section{Evolving Budgeted, Sequential Classifiers} \label{sec:emsco}
    This section details the construction and behavior of \texttt{EMSCO}. We first describe each of its components independently and then connect them for summary in Subsection \ref{sec:algorithm}. 
 
    \subsection{Chromosome Representation}\label{representation}
    To employ an evolutionary approach a chromosome representation for BSC designs  (ordered feature set partitions) is needed. 
    Ordered feature set partitions can be conveniently represented as lists of integers. Let $Q$ be a solution in $\mathcal{S}_{(n,k)}$. Then the chromosomal representation of $Q$ is denoted as $[Q]$, where 
    \begin{equation}
    [Q] \in \{0,1,\ldots,k-1\}^n.
    \end{equation} 
    The $j$\textsuperscript{th} list element $\Large{[}Q\Large{]}_j$ corresponds to the $j$th feature's stage assignment. The elements in $\Large{[}Q\Large{]}$ are \textit{zero-indexed}; so $\Large{[}Q\Large{]}_j = s$ assigns feature $j$ to stage $s+1$. For instance, $\{0,0,2,1,2\}$ represents a  system in which the first two features are assigned to the first stage, the fourth feature is assigned to the second stage, and the third and fifth features are assigned to the third stage. Note, stage count of solution $\Large{}Q\Large{}$ is denoted as $|Q|$, with $|Q| = (\max \Large{[}Q\Large{]}) + 1$.

     This relation between chromosome representations and ordered feature set partitions is not bijective. For example, $\{0,0,0,2\} \in \{0,1,2\}^4$, but since stage two is empty, this chromosome does not correspond to any ordered feature set partition in $\mathcal{S}_{(4,3)}$.
    This issue is attenuating for $n$ large relative to $k$, but as a proactive remedy, \texttt{EMSCO} checks for empty stages with a function, $gaps(\Large{[}Q\Large{]})$, that returns true if any stage is empty. If any such gaps exist, the solution is modified using a ``stage compression'' procedure (Algorithm \ref{alg:compression}).
\begin{algorithm}[h]
    \begin{footnotesize}
    \SetAlgoLined
    \textbf{function} $compress(\Large{[}Q\Large{]})$\\ {
    \If{$gaps(\Large{[}Q\Large{]})$}{
        $current\_stage \gets 0$;\\
        \tcp{$unique(S)$ returns unique elements in S.}
        \For{$stage \in unique(\Large{[}Q\Large{]})$} {
        \For{$index,value \in enumerate(\Large{[}Q\Large{]})$} {
        
        \If{$value = stage$} {
        $\Large{[}Q\Large{]}_{index} \gets current\_stage$;\\
        }
        $current\_stage \gets current\_stage + 1;$\\
        }
        }}
    \Return{$\Large{[}Q\Large{]}$};
    }
    \caption{Stage Compression}
    \label{alg:compression}
    \end{footnotesize}
    \end{algorithm}
For instance, $\{0,0,2,3\}$ is compressed to $\{0,0,1,2\}$ in the following steps: $$\{0,0,2,3\} \rightarrow \{0,0,1,3\} \rightarrow \{0,0,1,2\}.$$
    \subsection{Evolutionary Processes}\label{sec:operators}
    \texttt{EMSCO}'s evolutionary operators are designed to leverage problem-specific knowledge while maintaining sufficient breadth of search. These operators are applied in sequence to form a new population per generation. Each operator is detailed in the following subsections.
\subsubsection{Selection} \label{sec:selection}
Selection is the process in which individual solutions (chromosomes) are chosen from a population for the recombination stage. \texttt{EMSCO} utilizes roulette wheel selection (RWS) \cite{eiben}. In this scheme, a chromosome's probability of selection is proportional to its \textit{fitness} (as described in Section \ref{sec:fitness}). That is, solutions with greater fitness have a greater probability of being selected while not completely excluding solutions with low fitness. This latter property is considered desirable in our problem context. 
\subsubsection{Elitism} \label{sec:elitism}
The elitism protocol applied by \texttt{EMSCO} is designed to preserve all \textit{unique} non-dominated solutions to the subsequent generation. Let $E_0^*$ denote the set of unique non-dominated chromosomes within generation $G_h$, and let $G_h^*$ denote the set of all unique solutions in generation $G_h$. 
For elitism parameter $b\in [0,1]$, the top $$M = \max\left(round\left( b\cdot|G_h^*|\right), |E_0^*|\right),$$ unique solutions according to fitness (\ref{formula:fitness}) in each generation are sent to the subsequent generation without modification. Note, $round(x)$ returns the nearest integer for $x \in \mathbb{R}^+$. As can be seen in Algorithm \ref{alg:emsco}, this elitism protocol requires only $|G| - M \leq |G|$ chromosomes to be created and evaluated in each generation. 
\subsubsection{Mutation}\label{sec:mutation}

In general, mutation operators are used to maintain genetic diversity from one generation's population of chromosomes to the next \cite{eiben}. In addition to maintaing diversity, we design \texttt{EMSCO}'s mutation operator to leverage problem-specific aspects of sequential, budgeted classification.

If performance is comparable between solutions $Q_A,Q_B \in G_h$, the solution with lower stage count is preferred since it is more appealing from the perspective of model simplicity and its worst-case number of classifier evaluations is lower than the alternative.
Thus, to encourage lower stage counts during optimization, a discrete, monotonically decreasing probability distribution is desired to mutate chromosomes' stage assignments. The beta-binomial distribution with $\alpha =1 \And \beta > \alpha$, satisfies these criteria and possesses several convenient properties for adjusting bias of the mutation operator toward low stage assignments. 

Let $0 \leq j \leq |Q|$ be a potential stage assignment for the $i$\textsuperscript{th} feature in chromosome $Q$. At index $i$, if $rand(0,1) < \hat{m}$, the probability mass function for stage assignment $[Q]_i$ is:
\begin{equation} \label{eqn:mutation3}
    \mathbb{P}_{_{{\tiny{mut.}}}}\left ( \Large{[}Q\Large{]}_i \gets j \right ) = \binom{|Q|}{j} \frac{B\left (j + 1, |Q|-j+\beta\right )}{B\left (1, \beta\right)}
\end{equation}

where $B(\cdot)$ denotes the beta function. For a chromosome with stage count $|Q|$, this distribution has expected value $\frac{|Q|}{\beta + 1}$, and $\beta$ can be increased (decreased) to increase (decrease) bias towards low stage counts.

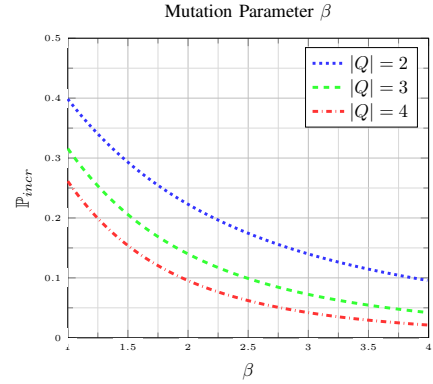
\begin{figure}
\centering
\begin{tikzpicture}[scale=\graphScale]

\begin{axis}[grid style={line width=.15pt, draw=gray!30},xmin=1,xmax=4,ymax=.5,ymin=0,major grid style={line width=.1pt,draw=gray!50},title={Mutation Parameter $\beta$},ylabel={$\mathbb{P}_{incr}$},xlabel=$\beta$,legend pos=north east, grid = both,
    minor x tick num={1},
    minor y tick num={1},
    minor z tick num={1},
    enlargelimits={abs=0},
    ticklabel style={font=\tiny,fill=white},
    axis line style={latex-latex}]

\addplot[
color=blue!80,
dotted,
ultra thick]
coordinates {
(1.0,0.3986169993576385)
(1.01,0.39603600017063345)
(1.02,0.39347720308274803)
(1.03,0.3909404052205847)
(1.04,0.3884254047273906)
(1.05,0.3859320008055611)
(1.06,0.38345999375640594)
(1.07,0.3810091850172985)
(1.08,0.3785793771963234)
(1.09,0.37617037410454124)
(1.1,0.3737819807859727)
(1.11,0.37141400354541176)
(1.12,0.36906624997416426)
(1.13,0.36673852897379755)
(1.1400000000000001,0.3644306507780156)
(1.15,0.36214242697272037)
(1.16,0.35987367051436914)
(1.17,0.3576241957466807)
(1.18,0.35539381841579276)
(1.19,0.353182355683929)
(1.2,0.35098962614165286)
(1.21,0.34881544981876944)
(1.22,0.3466596481939511)
(1.23,0.3445220442031385)
(1.24,0.3424024622467787)
(1.25,0.3403007281959667)
(1.26,0.3382166693975279)
(1.27,0.33615011467811096)
(1.28,0.33410089434733337)
(1.29,0.33206884020001937)
(1.3,0.33005378551759634)
(1.31,0.32805556506867206)
(1.32,0.32607401510885026)
(1.33,0.32410897337981825)
(1.34,0.3221602791077427)
(1.35,0.3202277730010157)
(1.3599999999999999,0.31831129724738805)
(1.37,0.31641069551051004)
(1.38,0.31452581292592974)
(1.3900000000000001,0.3126564960965724)
(1.4,0.3108025930877213)
(1.4100000000000001,0.3089639534215399)
(1.42,0.30714042807116826)
(1.43,0.30533186945439006)
(1.44,0.30353813142693864)
(1.45,0.30175906927541984)
(1.46,0.2999945397099131)
(1.47,0.29824440085624004)
(1.48,0.29650851224795394)
(1.49,0.2947867348180345)
(1.5,0.29307893089033654)
(1.51,0.29138496417078985)
(1.52,0.28970469973838564)
(1.53,0.28803800403594293)
(1.54,0.28638474486069443)
(1.55,0.2847447913546868)
(1.56,0.2831180139950279)
(1.57,0.28150428458397736)
(1.58,0.27990347623889866)
(1.5899999999999999,0.2783154633821)
(1.6,0.27674012173054563)
(1.6099999999999999,0.27517732828548025)
(1.62,0.27362696132195374)
(1.63,0.2720889003782707)
(1.6400000000000001,0.2705630262453679)
(1.65,0.26904922095612294)
(1.6600000000000001,0.26754736777462407)
(1.67,0.26605735118537843)
(1.6800000000000002,0.2645790568824975)
(1.69,0.2631123717588477)
(1.7000000000000002,0.26165718389516934)
(1.71,0.2602133825491998)
(1.72,0.25878085814476937)
(1.73,0.25735950226089777)
(1.74,0.2559492076209059)
(1.75,0.2545498680815116)
(1.76,0.25316137862196353)
(1.77,0.2517836353331784)
(1.78,0.2504165354068969)
(1.79,0.24905997712488626)
(1.8,0.24771385984814853)
(1.81,0.24637808400618388)
(1.82,0.24505255108628587)
(1.83,0.24373716362286735)
(1.8399999999999999,0.2424318251868547)
(1.85,0.24113644037511062)
(1.8599999999999999,0.23985091479991638)
(1.87,0.23857515507850557)
(1.88,0.23730906882266)
(1.8900000000000001,0.23605256462835122)
(1.9,0.23480555206546194)
(1.9100000000000001,0.23356794166755246)
(1.92,0.2323396449217029)
(1.9300000000000002,0.23112057425841814)
(1.94,0.22991064304160302)
(1.9500000000000002,0.2287097655586049)
(1.96,0.22751785701033245)
(1.97,0.22633483350144112)
(1.98,0.225160612030596)
(1.99,0.2239951104808091)
(2.0,0.2228382476098566)
(2.01,0.22168994304075917)
(2.02,0.22055011725236306)
(2.0300000000000002,0.2194186915699754)
(2.04,0.21829558815609718)
(2.05,0.2171807300012223)
(2.06,0.21607404091473148)
(2.0700000000000003,0.21497544551585135)
(2.08,0.21388486922470462)
(2.09,0.21280223825343936)
(2.1,0.21172747959743843)
(2.1100000000000003,0.21066052102661315)
(2.12,0.2096012910767786)
(2.13,0.20854971904110564)
(2.14,0.20750573496166602)
(2.1500000000000004,0.2064692696210476)
(2.16,0.20544025453406234)
(2.17,0.20441862193952787)
(2.1799999999999997,0.20340430479213312)
(2.19,0.20239723675439503)
(2.2,0.20139735218867827)
(2.21,0.200404586149317)
(2.2199999999999998,0.19941887437480033)
(2.23,0.1984401532800505)
(2.24,0.19746835994877454)
(2.25,0.19650343212590193)
(2.26,0.19554530821009597)
(2.27,0.1945939272463486)
(2.2800000000000002,0.19364922891865421)
(2.29,0.19271115354275847)
(2.3,0.19177964205899167)
(2.31,0.1908546360251755)
(2.3200000000000003,0.18993607760960363)
(2.33,0.18902390958410775)
(2.34,0.18811807531719438)
(2.35,0.18721851876725493)
(2.3600000000000003,0.186325184475858)
(2.37,0.18543801756111333)
(2.38,0.18455696371110408)
(2.39,0.18368196917740276)
(2.4000000000000004,0.1828129807686525)
(2.41,0.18194994584422208)
(2.42,0.18109281230793817)
(2.4299999999999997,0.18024152860187959)
(2.44,0.17939604370025264)
(2.45,0.17855630710332382)
(2.46,0.1777222688314325)
(2.4699999999999998,0.17689387941907297)
(2.48,0.17607108990903075)
(2.49,0.1752538518466057)
(2.5,0.17444211727388415)
(2.51,0.1736358387240935)
(2.52,0.17283496921600738)
(2.5300000000000002,0.17203946224842637)
(2.54,0.1712492717947195)
(2.55,0.17046435229742696)
(2.56,0.1696846586629338)
(2.5700000000000003,0.16891014625619427)
(2.58,0.16814077089552726)
(2.59,0.16737648884747414)
(2.6,0.1666172568217047)
(2.6100000000000003,0.16586303196599717)
(2.62,0.1651137718612724)
(2.63,0.1643694345166834)
(2.64,0.16362997836476734)
(2.6500000000000004,0.16289536225665158)
(2.66,0.16216554545731854)
(2.67,0.16144048764093166)
(2.6799999999999997,0.160720148886201)
(2.69,0.16000448967182823)
(2.7,0.1592934708719801)
(2.71,0.15858705375183224)
(2.7199999999999998,0.15788519996316852)
(2.73,0.1571878715400158)
(2.74,0.15649503089434513)
(2.75,0.15580664081182494)
(2.76,0.15512266444761358)
(2.77,0.1544430653222103)
(2.7800000000000002,0.15376780731736062)
(2.79,0.15309685467199707)
(2.8,0.15243017197824094)
(2.81,0.15176772417744488)
(2.8200000000000003,0.15110947655628282)
(2.83,0.15045539474289749)
(2.84,0.14980544470307544)
(2.85,0.14915959273648738)
(2.8600000000000003,0.14851780547295412)
(2.87,0.14788004986877779)
(2.88,0.14724629320310367)
(2.89,0.14661650307432528)
(2.9000000000000004,0.14599064739653844)
(2.91,0.1453686943960395)
(2.92,0.14475061260785516)
(2.9299999999999997,0.14413637087232745)
(2.94,0.14352593833172733)
(2.95,0.14291928442691604)
(2.96,0.1423163788940468)
(2.9699999999999998,0.14171719176130126)
(2.98,0.1411216933456717)
(2.99,0.14052985424977582)
(3.0,0.13994164535871156)
(3.0100000000000002,0.13935703783695064)
(3.02,0.13877600312526872)
(3.0300000000000002,0.13819851293771457)
(3.04,0.13762453925861273)
(3.05,0.13705405433959983)
(3.06,0.1364870306967022)
(3.07,0.13592344110744603)
(3.08,0.13536325860799903)
(3.09,0.13480645649035183)
(3.1,0.13425300829952724)
(3.11,0.13370288783083284)
(3.12,0.1331560691271262)
(3.13,0.13261252647614286)
(3.14,0.13207223440782245)
(3.15,0.1315351676916965)
(3.16,0.13100130133429022)
(3.17,0.13047061057655485)
(3.18,0.1299430708913435)
(3.19,0.1294186579809007)
(3.2,0.128897347774397)
(3.21,0.12837911642548128)
(3.22,0.12786394030986725)
(3.23,0.1273517960229512)
(3.24,0.12684266037745107)
(3.25,0.12633651040107619)
(3.2600000000000002,0.12583332333423425)
(3.27,0.12533307662774595)
(3.2800000000000002,0.12483574794060515)
(3.29,0.12434131513775448)
(3.3000000000000003,0.12384975628789663)
(3.31,0.12336104966131622)
(3.32,0.12287517372774515)
(3.33,0.1223921071542391)
(3.34,0.12191182880309381)
(3.35,0.12143431772976387)
(3.36,0.12095955318082918)
(3.37,0.12048751459197593)
(3.38,0.12001818158599131)
(3.39,0.11955153397080043)
(3.4,0.1190875517375134)
(3.41,0.11862621505849702)
(3.42,0.11816750428547573)
(3.43,0.1177113999476449)
(3.44,0.11725788274981641)
(3.45,0.11680693357057537)
(3.46,0.11635853346046898)
(3.47,0.11591266364020558)
(3.48,0.11546930549888956)
(3.49,0.11502844059225625)
(3.5,0.11459005064094907)
(3.5100000000000002,0.11415411752879756)
(3.52,0.11372062330113408)
(3.5300000000000002,0.11328955016311049)
(3.54,0.11286088047805443)
(3.5500000000000003,0.11243459676582113)
(3.56,0.1120106817011931)
(3.57,0.11158911811226901)
(3.58,0.11116988897889535)
(3.59,0.11075297743109913)
(3.6,0.11033836674754693)
(3.61,0.10992604035402376)
(3.62,0.10951598182191957)
(3.63,0.10910817486674529)
(3.64,0.10870260334665671)
(3.65,0.10829925126099915)
(3.66,0.10789810274886669)
(3.67,0.10749914208768618)
(3.68,0.10710235369180088)
(3.69,0.10670772211108714)
(3.7,0.10631523202957804)
(3.71,0.10592486826410552)
(3.72,0.10553661576295437)
(3.73,0.10515045960453795)
(3.74,0.10476638499607949)
(3.75,0.10438437727232286)
(3.7600000000000002,0.1040044218942413)
(3.77,0.10362650444777421)
(3.7800000000000002,0.10325061064256746)
(3.79,0.10287672631073785)
(3.8000000000000003,0.10250483740564853)
(3.81,0.10213493000068785)
(3.82,0.10176699028808145)
(3.83,0.10140100457770074)
(3.84,0.10103695929589118)
(3.85,0.10067484098431934)
(3.86,0.10031463629881943)
(3.87,0.09995633200826748)
(3.88,0.09959991499345533)
(3.89,0.09924537224598984)
(3.9,0.09889269086719155)
(3.91,0.0985418580670151)
(3.92,0.09819286116298132)
(3.93,0.09784568757910916)
(3.94,0.09750032484487936)
(3.95,0.09715676059419298)
(3.96,0.09681498256434629)
(3.97,0.0964749785950253)
(3.98,0.09613673662729527)
(3.99,0.09580024470262005)
};

\addplot[
color=green!80,
dashed,
ultra thick]
coordinates {
(1.0,0.3159793144360765)
(1.01,0.3131446400907816)
(1.02,0.31034056385351627)
(1.03,0.3075667489791807)
(1.04,0.3048228609529947)
(1.05,0.3021085675627818)
(1.06,0.2994235389657862)
(1.07,0.2967674477502805)
(1.08,0.29413996899222405)
(1.09,0.29154078030719976)
(1.1,0.2889695618978906)
(1.11,0.28642599659726353)
(1.12,0.28390976990773964)
(1.13,0.28142057003648935)
(1.1400000000000001,0.27895808792708765)
(1.15,0.27652201728770154)
(1.16,0.27411205461598076)
(1.17,0.27172789922082785)
(1.18,0.2693692532412132)
(1.19,0.267035821662178)
(1.2,0.2647273123281806)
(1.21,0.2624434359539327)
(1.22,0.26018390613284237)
(1.23,0.25794843934321277)
(1.24,0.2557367549523095)
(1.25,0.25354857521841756)
(1.26,0.2513836252909877)
(1.27,0.2492416332090105)
(1.28,0.2471223298976759)
(1.29,0.24502544916345992)
(1.3,0.24295072768769732)
(1.31,0.24089790501875608)
(1.32,0.23886672356287408)
(1.33,0.2368569285737545)
(1.34,0.23486826814100015)
(1.35,0.2329004931774401)
(1.3599999999999999,0.23095335740544265)
(1.37,0.22902661734226393)
(1.38,0.22712003228450528)
(1.3900000000000001,0.2252333642917308)
(1.4,0.22336637816932114)
(1.4100000000000001,0.22151884145058998)
(1.42,0.21969052437824443)
(1.43,0.2178811998852178)
(1.44,0.216090643574938)
(1.45,0.21431863370105597)
(1.46,0.21256495114670904)
(1.47,0.2108293794033178)
(1.48,0.20911170454900074)
(1.49,0.20741171522660384)
(1.5,0.20572920262140382)
(1.51,0.20406396043850816)
(1.52,0.20241578487999645)
(1.53,0.2007844746218006)
(1.54,0.19916983079040307)
(1.55,0.19757165693932677)
(1.56,0.1959897590254781)
(1.57,0.19442394538535512)
(1.58,0.19287402671114307)
(1.5899999999999999,0.19133981602671524)
(1.6,0.1898211286635738)
(1.6099999999999999,0.18831778223672635)
(1.62,0.1868295966205491)
(1.63,0.18535639392461567)
(1.6400000000000001,0.18389799846953736)
(1.65,0.18245423676282424)
(1.6600000000000001,0.18102493747476112)
(1.67,0.17960993141435122)
(1.6800000000000002,0.17820905150529098)
(1.69,0.17682213276203562)
(1.7000000000000002,0.17544901226593845)
(1.71,0.17408952914147247)
(1.72,0.17274352453256925)
(1.73,0.1714108415790534)
(1.74,0.17009132539320793)
(1.75,0.16878482303645737)
(1.76,0.16749118349619363)
(1.77,0.1662102576627379)
(1.78,0.16494189830645423)
(1.79,0.16368596005501745)
(1.8,0.16244229937083798)
(1.81,0.16121077452866017)
(1.82,0.1599912455933239)
(1.83,0.1587835743976972)
(1.8399999999999999,0.15758762452080144)
(1.85,0.15640326126610649)
(1.8599999999999999,0.15523035164001453)
(1.87,0.15406876433053862)
(1.88,0.15291836968616757)
(1.8900000000000001,0.15177903969492446)
(1.9,0.1506506479636256)
(1.9100000000000001,0.14953306969733415)
(1.92,0.1484261816790159)
(1.9300000000000002,0.1473298622493988)
(1.94,0.14624399128702859)
(1.9500000000000002,0.14516845018853763)
(1.96,0.14410312184911322)
(1.97,0.14304789064316614)
(1.98,0.14200264240522187)
(1.99,0.14096726441100071)
(2.0,0.13994164535871156)
(2.01,0.1389256753505569)
(2.02,0.13791924587443782)
(2.0300000000000002,0.13692224978586798)
(2.04,0.13593458129009584)
(2.05,0.13495613592442768)
(2.06,0.13398681054075945)
(2.0700000000000003,0.13302650328831456)
(2.08,0.13207511359658086)
(2.09,0.1311325421584557)
(2.1,0.1301986909135845)
(2.1100000000000003,0.12927346303191123)
(2.12,0.1283567628974227)
(2.13,0.12744849609209352)
(2.14,0.12654856938001924)
(2.1500000000000004,0.12565689069176667)
(2.16,0.12477336910890058)
(2.17,0.12389791484870605)
(2.1799999999999997,0.12303043924911183)
(2.19,0.12217085475380163)
(2.2,0.12131907489750415)
(2.21,0.12047501429148988)
(2.2199999999999998,0.11963858860923127)
(2.23,0.11880971457227107)
(2.24,0.11798830993625076)
(2.25,0.11717429347713804)
(2.26,0.11636758497762756)
(2.27,0.11556810521371075)
(2.2800000000000002,0.11477577594143751)
(2.29,0.11399051988384135)
(2.3,0.1132122607180387)
(2.31,0.11244092306250297)
(2.3200000000000003,0.1116764324645021)
(2.33,0.11091871538770615)
(2.34,0.11016769919996638)
(2.35,0.10942331216124124)
(2.3600000000000003,0.10868548341170381)
(2.37,0.10795414295999839)
(2.38,0.10722922167165128)
(2.39,0.10651065125765424)
(2.4000000000000004,0.10579836426318256)
(2.41,0.1050922940564809)
(2.42,0.10439237481789498)
(2.4299999999999997,0.10369854152905156)
(2.44,0.10301072996218341)
(2.45,0.10232887666960999)
(2.46,0.10165291897335227)
(2.4699999999999998,0.10098279495488893)
(2.48,0.10031844344506413)
(2.49,0.09965980401412156)
(2.5,0.09900681696187918)
(2.51,0.09835942330804914)
(2.52,0.09771756478267324)
(2.5300000000000002,0.09708118381670072)
(2.54,0.09645022353270216)
(2.55,0.09582462773569633)
(2.56,0.09520434090411367)
(2.5700000000000003,0.09458930818088696)
(2.58,0.09397947536465823)
(2.59,0.09337478890111373)
(2.6,0.0927751958744314)
(2.6100000000000003,0.09218064399885462)
(2.62,0.0915910816103862)
(2.63,0.09100645765857962)
(2.64,0.0904267216984671)
(2.6500000000000004,0.08985182388258561)
(2.66,0.08928171495311688)
(2.67,0.08871634623413904)
(2.6799999999999997,0.08815566962398236)
(2.69,0.08759963758769995)
(2.7,0.0870482031496268)
(2.71,0.08650131988606335)
(2.7199999999999998,0.08595894191804387)
(2.73,0.08542102390421613)
(2.74,0.08488752103381247)
(2.75,0.08435838901972692)
(2.76,0.08383358409167996)
(2.77,0.08331306298948937)
(2.7800000000000002,0.08279678295642245)
(2.79,0.08228470173265368)
(2.8,0.08177677754880164)
(2.81,0.08127296911956172)
(2.8200000000000003,0.08077323563743344)
(2.83,0.08027753676652327)
(2.84,0.07978583263643602)
(2.85,0.07929808383626968)
(2.8600000000000003,0.07881425140865894)
(2.87,0.07833429684393778)
(2.88,0.07785818207435946)
(2.89,0.07738586946840675)
(2.9000000000000004,0.07691732182517952)
(2.91,0.07645250236885537)
(2.92,0.07599137474324036)
(2.9299999999999997,0.07553390300637752)
(2.94,0.07508005162524856)
(2.95,0.07462978547053323)
(2.96,0.07418306981145051)
(2.9699999999999998,0.07373987031066964)
(2.98,0.0733001530192916)
(2.99,0.07286388437189817)
(3.0,0.07243103118167216)
(3.0100000000000002,0.07200156063558327)
(3.02,0.07157544028964058)
(3.0300000000000002,0.07115263806421379)
(3.04,0.07073312223941497)
(3.05,0.07031686145054539)
(3.06,0.06990382468360223)
(3.07,0.06949398127086204)
(3.08,0.06908730088649728)
(3.09,0.06868375354228384)
(3.1,0.06828330958334228)
(3.11,0.06788593968395784)
(3.12,0.06749161484344235)
(3.13,0.06710030638205788)
(3.14,0.06671198593700645)
(3.15,0.06632662545845547)
(3.16,0.06594419720563438)
(3.17,0.06556467374297748)
(3.18,0.06518802793632028)
(3.19,0.06481423294914401)
(3.2,0.06444326223888264)
(3.21,0.06407508955326857)
(3.22,0.06370968892673068)
(3.23,0.06334703467685177)
(3.24,0.06298710140085484)
(3.25,0.06262986397215498)
(3.2600000000000002,0.062275297536951935)
(3.27,0.0619233775108633)
(3.2800000000000002,0.061574079575615226)
(3.29,0.061227379675761084)
(3.3000000000000003,0.06088325401546735)
(3.31,0.06054167905532026)
(3.32,0.060202631509188365)
(3.33,0.0598660883411285)
(3.34,0.0595320267623225)
(3.35,0.059200424228068704)
(3.36,0.05887125843480412)
(3.37,0.058544507317171535)
(3.38,0.05822014904512707)
(3.39,0.05789816202107545)
(3.4,0.05757852487706461)
(3.41,0.05726121647199556)
(3.42,0.05694621588888715)
(3.43,0.0566335024321607)
(3.44,0.056323055624982965)
(3.45,0.05601485520662153)
(3.46,0.05570888112984729)
(3.47,0.0554051135583743)
(3.48,0.055103532864326565)
(3.49,0.05480411962574083)
(3.5,0.05450685462410376)
(3.5100000000000002,0.054211718841920886)
(3.52,0.05391869346031808)
(3.5300000000000002,0.05362775985667356)
(3.54,0.053338899602286216)
(3.5500000000000003,0.05305209446006154)
(3.56,0.052767326382251456)
(3.57,0.05248457750819413)
(3.58,0.052203830162108944)
(3.59,0.05192506685091358)
(3.6,0.05164827026205343)
(3.61,0.05137342326138794)
(3.62,0.0511005088910802)
(3.63,0.0508295103675267)
(3.64,0.05056041107931053)
(3.65,0.05029319458518344)
(3.66,0.05002784461207088)
(3.67,0.04976434505310634)
(3.68,0.04950267996568736)
(3.69,0.049242833569561606)
(3.7,0.04898479024493685)
(3.71,0.048728534530608436)
(3.72,0.048474051122121)
(3.73,0.048221324869951454)
(3.74,0.04797034077770812)
(3.75,0.047721084000364455)
(3.7600000000000002,0.04747353984250546)
(3.77,0.04722769375659985)
(3.7800000000000002,0.046983531341305285)
(3.79,0.046741038339770946)
(3.8000000000000003,0.046500200637990874)
(3.81,0.046261004263155714)
(3.82,0.046023435382033906)
(3.83,0.0457874802993814)
(3.84,0.04555312545635315)
(3.85,0.045320357428958236)
(3.86,0.04508916292651144)
(3.87,0.044859528790121006)
(3.88,0.04463144199118707)
(3.89,0.04440488962992861)
(3.9,0.04417985893391374)
(3.91,0.04395633725662407)
(3.92,0.043734312076027404)
(3.93,0.043513770993170775)
(3.94,0.04329470173080108)
(3.95,0.04307709213198008)
(3.96,0.042860930158741706)
(3.97,0.042646203890753465)
(3.98,0.04243290152399404)
(3.99,0.04222101136945511)
};

\addplot[
color=red!80,
dashdotted,
ultra thick]
coordinates {
(1.0,0.26143089735459624)
(1.01,0.2585478361906728)
(1.02,0.25570137988265174)
(1.03,0.2528910578796426)
(1.04,0.25011640382920297)
(1.05,0.24737695565597762)
(1.06,0.24467225563241413)
(1.07,0.24200185044199718)
(1.08,0.2393652912353864)
(1.09,0.23676213367985177)
(1.1,0.23419193800236404)
(1.11,0.23165426902670738)
(1.12,0.22914869620491396)
(1.13,0.22667479364337761)
(1.1400000000000001,0.2242321401239139)
(1.15,0.22182031912008005)
(1.16,0.21943891880901656)
(1.17,0.21708753207907638)
(1.18,0.21476575653348506)
(1.19,0.21247319449028934)
(1.2,0.2102094529787819)
(1.21,0.20797414373266798)
(1.22,0.2057668831801257)
(1.23,0.20358729243100238)
(1.24,0.201434997261293)
(1.25,0.1993096280951061)
(1.26,0.19721081998426848)
(1.27,0.19513821258572506)
(1.28,0.1930914501369102)
(1.29,0.1910701814291943)
(1.3,0.18907405977957614)
(1.31,0.18710274300073715)
(1.32,0.18515589336957838)
(1.33,0.1832331775943603)
(1.34,0.18133426678055153)
(1.35,0.17945883639550553)
(1.3599999999999999,0.17760656623204052)
(1.37,0.17577714037103742)
(1.38,0.17397024714314124)
(1.3900000000000001,0.17218557908963983)
(1.4,0.17042283292260907)
(1.4100000000000001,0.16868170948439964)
(1.42,0.1669619137065349)
(1.43,0.16526315456808305)
(1.44,0.16358514505358335)
(1.45,0.16192760211055957)
(1.46,0.1602902466067121)
(1.47,0.15867280328680233)
(1.48,0.15707500072932123)
(1.49,0.15549657130295835)
(1.5,0.15393725112292855)
(1.51,0.15239678000720325)
(1.52,0.1508749014326819)
(1.53,0.14937136249132188)
(1.54,0.1478859138463089)
(1.55,0.14641830968824765)
(1.56,0.14496830769143654)
(1.57,0.14353566897025072)
(1.58,0.1421201580356506)
(1.5899999999999999,0.14072154275184434)
(1.6,0.13933959429314136)
(1.6099999999999999,0.13797408710099401)
(1.62,0.13662479884127854)
(1.63,0.13529151036179876)
(1.6400000000000001,0.1339740056500629)
(1.65,0.13267207179133356)
(1.6600000000000001,0.13138549892696672)
(1.67,0.13011408021305426)
(1.6800000000000002,0.12885761177938837)
(1.69,0.12761589268875384)
(1.7000000000000002,0.12638872489656394)
(1.71,0.12517591321083954)
(1.72,0.12397726525255626)
(1.73,0.12279259141635634)
(1.74,0.12162170483163348)
(1.75,0.12046442132400437)
(1.76,0.11932055937716712)
(1.77,0.1181899400951476)
(1.78,0.11707238716494983)
(1.79,0.11596772681960854)
(1.8,0.11487578780164531)
(1.81,0.11379640132692737)
(1.82,0.11272940104895424)
(1.83,0.1116746230235327)
(1.8399999999999999,0.1106319056738887)
(1.85,0.10960108975618366)
(1.8599999999999999,0.10858201832544845)
(1.87,0.10757453670193506)
(1.88,0.10657849243788364)
(1.8900000000000001,0.10559373528470661)
(1.9,0.10462011716059039)
(1.9100000000000001,0.10365749211850639)
(1.92,0.1027057163146411)
(1.9300000000000002,0.1017646479772385)
(1.94,0.10083414737584429)
(1.9500000000000002,0.09991407679096942)
(1.96,0.09900430048415498)
(1.97,0.09810468466844136)
(1.98,0.09721509747923951)
(1.99,0.09633540894559844)
(2.0,0.09546549096187629)
(2.01,0.09460521725979532)
(2.02,0.09375446338089743)
(2.0300000000000002,0.09291310664937547)
(2.04,0.09208102614529934)
(2.05,0.09125810267821721)
(2.06,0.09044421876112618)
(2.0700000000000003,0.08963925858483879)
(2.08,0.0888431079926928)
(2.09,0.08805565445564201)
(2.1,0.08727678704771458)
(2.1100000000000003,0.0865063964218099)
(2.12,0.08574437478587593)
(2.13,0.08499061587941481)
(2.14,0.0842450149503492)
(2.1500000000000004,0.08350746873222759)
(2.16,0.0827778754217674)
(2.17,0.08205613465673922)
(2.1799999999999997,0.08134214749417001)
(2.19,0.08063581638888861)
(2.2,0.07993704517238165)
(2.21,0.07924573903197651)
(2.2199999999999998,0.07856180449033923)
(2.23,0.07788514938527358)
(2.24,0.07721568284984026)
(2.25,0.07655331529277343)
(2.26,0.07589795837919233)
(2.27,0.07524952501161664)
(2.2800000000000002,0.07460792931126359)
(2.29,0.07397308659964041)
(2.3,0.07334491338041549)
(2.31,0.07272332732156517)
(2.3200000000000003,0.07210824723781106)
(2.33,0.07149959307330356)
(2.34,0.07089728588460265)
(2.35,0.07030124782389924)
(2.3600000000000003,0.06971140212250415)
(2.37,0.06912767307460288)
(2.38,0.06854998602124718)
(2.39,0.0679782673346071)
(2.4000000000000004,0.0674124444024633)
(2.41,0.0668524456129439)
(2.42,0.0662982003394974)
(2.4299999999999997,0.06574963892610453)
(2.44,0.06520669267270951)
(2.45,0.06466929382089925)
(2.46,0.06413737553978571)
(2.4699999999999998,0.06361087191212311)
(2.48,0.063089717920638)
(2.49,0.06257384943457178)
(2.5,0.062063203196442984)
(2.51,0.06155771680899891)
(2.52,0.061057328722397775)
(2.5300000000000002,0.06056197822156584)
(2.54,0.06007160541377654)
(2.55,0.05958615121640165)
(2.56,0.05910555734487688)
(2.5700000000000003,0.05862976630084582)
(2.58,0.05815872136047928)
(2.59,0.05769236656300847)
(2.6,0.05723064669939937)
(2.6100000000000003,0.05677350730124453)
(2.62,0.05632089462979761)
(2.63,0.05587275566519434)
(2.64,0.05542903809584743)
(2.6500000000000004,0.05498969030799716)
(2.66,0.054554661375431523)
(2.67,0.054123901049368706)
(2.6799999999999997,0.053697359748496765)
(2.69,0.0532749885491699)
(2.7,0.052856739175755285)
(2.71,0.05244256399114189)
(2.7199999999999998,0.052032415987383285)
(2.73,0.05162624877649957)
(2.74,0.05122401658141906)
(2.75,0.05082567422705797)
(2.76,0.05043117713154277)
(2.77,0.050040481297577655)
(2.7800000000000002,0.0496535433039339)
(2.79,0.04927032029707701)
(2.8,0.048890769982931515)
(2.81,0.048514850618768146)
(2.8200000000000003,0.04814252100520955)
(2.83,0.047773740478384896)
(2.84,0.047408468902179424)
(2.85,0.04704666666062429)
(2.8600000000000003,0.04668829465039015)
(2.87,0.046333314273414694)
(2.88,0.045981687429628604)
(2.89,0.045633376509802925)
(2.9000000000000004,0.04528834438851059)
(2.91,0.044946554417184426)
(2.92,0.044607970417295584)
(2.9299999999999997,0.04427255667363861)
(2.94,0.04394027792770283)
(2.95,0.04361109937117158)
(2.96,0.043284986639501644)
(2.9699999999999998,0.04296190580561099)
(2.98,0.04264182337366185)
(2.99,0.042324706272944934)
(3.0,0.04201052185184839)
(3.0100000000000002,0.04169923787192442)
(3.02,0.041390822502052727)
(3.0300000000000002,0.041085244312683566)
(3.04,0.04078247227017828)
(3.05,0.04048247573122832)
(3.06,0.04018522443736605)
(3.07,0.03989068850956312)
(3.08,0.039598838442895246)
(3.09,0.039309645101312074)
(3.1,0.03902307971246732)
(3.11,0.038739113862635066)
(3.12,0.03845771949170651)
(3.13,0.038178868888260054)
(3.14,0.03790253468470639)
(3.15,0.03762868985250645)
(3.16,0.03735730769746293)
(3.17,0.03708836185508724)
(3.18,0.036821826286029036)
(3.19,0.0365576752715816)
(3.2,0.03629588340925205)
(3.21,0.036036425608402034)
(3.22,0.03577927708594231)
(3.23,0.03552441336211598)
(3.24,0.03527181025631598)
(3.25,0.03502144388299033)
(3.2600000000000002,0.0347732906475986)
(3.27,0.03452732724262253)
(3.2800000000000002,0.034283530643652194)
(3.29,0.03404187810551429)
(3.3000000000000003,0.033802347158472745)
(3.31,0.033564915604478385)
(3.32,0.03332956151347344)
(3.33,0.033096263219753785)
(3.34,0.03286499931839304)
(3.35,0.03263574866170271)
(3.36,0.03240849035575666)
(3.37,0.03218320375697303)
(3.38,0.031959868468729336)
(3.39,0.03173846433804428)
(3.4,0.031518971452298206)
(3.41,0.03130137013600598)
(3.42,0.031085640947642657)
(3.43,0.03087176467650432)
(3.44,0.030659722339626883)
(3.45,0.03044949517873874)
(3.46,0.0302410646572715)
(3.47,0.03003441245740479)
(3.48,0.029829520477158145)
(3.49,0.029626370827523063)
(3.5,0.02942494582964139)
(3.5100000000000002,0.029225228012017324)
(3.52,0.029027200107785034)
(3.5300000000000002,0.02883084505199618)
(3.54,0.028636145978960803)
(3.5500000000000003,0.028443086219626657)
(3.56,0.028251649298985715)
(3.57,0.028061818933534988)
(3.58,0.027873579028756756)
(3.59,0.027686913676648883)
(3.6,0.02750180715328343)
(3.61,0.027318243916401808)
(3.62,0.027136208603046552)
(3.63,0.02695568602722942)
(3.64,0.02677666117762345)
(3.65,0.026599119215299116)
(3.66,0.02642304547148533)
(3.67,0.026248425445368095)
(3.68,0.02607524480192014)
(3.69,0.025903489369752286)
(3.7,0.025733145139012925)
(3.71,0.02556419825930012)
(3.72,0.02539663503761147)
(3.73,0.025230441936328618)
(3.74,0.02506560557121762)
(3.75,0.024902112709465407)
(3.7600000000000002,0.02473995026774678)
(3.77,0.024579105310315597)
(3.7800000000000002,0.024419565047115288)
(3.79,0.024261316831930668)
(3.8000000000000003,0.02410434816055651)
(3.81,0.023948646668994544)
(3.82,0.02379420013166822)
(3.83,0.02364099645968476)
(3.84,0.023489023699087563)
(3.85,0.02333827002917066)
(3.86,0.02318872376078207)
(3.87,0.02304037333467701)
(3.88,0.02289320731988076)
(3.89,0.022747214412077055)
(3.9,0.02260238343202381)
(3.91,0.02245870332397748)
(3.92,0.02231616315415763)
(3.93,0.022174752109221263)
(3.94,0.02203445949475613)
(3.95,0.021895274733804704)
(3.96,0.02175718736540455)
(3.97,0.021620187043138728)
(3.98,0.021484263533726344)
(3.99,0.021349406715612473)
};

\legend{$|Q|=2$,$|Q|=3$,$|Q|=4$}

\end{axis}

\end{tikzpicture}

\caption{Greater $\beta$ corresponds to lower probability of incrementing stage count. Fifteen features and a mutation rate of $10\%$ are assumed for this demonstration.}
\label{fig:mutation}
\end{figure}

 When warranted, it is important that the mutation operator allows creation of new stages---in scenarios with many expensive features, a low stage count may yield high feature acquisition cost at each stage. The mutation operator creates new stages in $\Large{[}Q\Large{]}$ by assigning $|Q|$ to an element\footnote{Recall, the list representation is zero-indexed at each element, so a stage assignment of $|Q|$ sends the feature to new stage $|Q| + 1$}. Let $$p_{_{Q^+}} = \mathbb{P}_{mut}\left([Q]_j \gets |Q| \right) = \beta \left ( \frac{\Gamma (|Q| +1) \cdot \Gamma (\beta)}{\Gamma \left(|Q| + \beta + 1\right)} \right )$$ For mutation rate $\hat{m}$ and number of features $n$, the probability that the mutation operator increments solution $Q$'s stage count is given by:
 \begin{equation}
     \mathbb{P}_{incr}(Q) = 1 - (1 - \hat{m}\cdot p_{_{Q^+}})^n,
 \end{equation}
 since this is the complement of the event in which no features are assigned stage $|Q|$. The effect of $\beta$ on $\mathbb{P}_{incr}$ is depicted in Figure \ref{fig:mutation}.
    \subsubsection{Recombination}
    \texttt{EMSCO}'s recombination/crossover operator takes into account the differing contexts of stage assignments within the parent solutions. To implement crossover, two parent chromosomes $P_A, P_B \in G_h$ are selected according to the protocol described in Section \ref{sec:selection}. With probability $1 - \hat{r}$, recombination/crossover sends one of $P_A,P_B$  to $G_{h+1}$ unchanged via a fair coin flip. With probability $\hat{r}$, $P_A$ and $P_B$ are combined with a problem-specific variant of uniform crossover \cite{eiben} to produce a child $C \in G_{h+1}$. In this case, the first step in recombination of $P_A$ and $P_B$ randomly selects a stage count for the child solution based on the parents' stage counts---with equal probabilities ($.3\bar{3})$, three options are considered for $|C|$: $|P_A|, |P_B|,round\left( \frac{|P_A|+|P_B|}{2}\right)$, where the third option is incorporated to facilitate consideration of solutions with intermediate stage counts between the parents'. After a stage count is decided, each of the child's features must be assigned a stage. Each $i$th assignment is initially chosen to be either $\Large{[}P_A\Large{]}_i$ or $\Large{[}P_B\Large{]}_i$ with equal probability ($0.5$). Let $R \in \{P_A, P_B\}$ denote the parent corresponding to this choice. In an attempt to maintain context of the parent's stage assignment for the $i$th feature, the initial selection $\Large{[}R\Large{]}_i$ is first normalized according to $|R|$ to acquire the relative order of the stage assignment within parent solution $R$. This ratio is then multiplied by $|C|$, rounded, and decremented to obtain a zero-indexed, approximate analog of $\Large{[}R\Large{]}_i$ in $\Large{[}C\Large{]}$.

    \subsection{Evaluating Chromosomes}\label{sec:fitness}

    For a solution $Q \in G_h$, fitness is computed using the $L_2$ norm of $\left\{g_1(Q),g_2(Q),g_3(Q)\right\}$, denoted by $\mathscr{E}(Q)$, \textit{and} an exponential term accounting for the \textit{Pareto rank} of $Q$ with respect to the current population. In this paper, Pareto rank, or \textit{non-domination level} \cite{deb02}, is an indication of the solution's performance from the perspective of Pareto efficiency. This mixed formulation of fitness respects Pareto efficiency and raw aggregated performance simultaneously. 
    \subsubsection{Pareto Rank of Chromosomes}
   To compute $rank(Q)$ in generation $G_h$, we first determine the non-dominated solutions in $G_h$.  This set of solutions $E_0$ is then removed from $G_h$. The non-dominated solutions in set $G_h \setminus E_0$ are then stored in the set $E_1$. This process of removing non-dominated solutions continues so that $E_t$ contains the non-dominated solutions in the population $G_h \setminus E_0 \setminus E_1 \setminus \ldots \setminus E_{t-1}$. Once all solutions have been assigned to some $E_t$, the ranking procedure is finished. 
    \
    Let $E_{t^*}$ denote the final non-dominated set removed from $G_h$, and let $Q \in E_t$. Then 
    \begin{equation}\label{rank}
    rank(Q) = t^* - t.
    \end{equation}
     Our assignment of non-domination level is flipped--- the first non-dominated set is commonly assigned zero rank \cite{deb02}. However, we define $rank(\cdot)$ so that the initial set of non-dominated solutions is assigned the greatest value. The motivation for this becomes clearer in the following subsection, where we define fitness.
    \subsubsection{Fitness Function} 
    Non-domination is an important aspect of performance, and as seen in (\ref{problem}), defines \texttt{EMSCO}'s optimization problem formally. However, in the context of BSC design, relying on non-domination level exclusively can lead to an ineffective prioritization of objectives during selection because there may be important practical differences among solutions in the same non-domination level. As a remedy, we consider rank \textit{and} scalarized performance simultaneously via the fitness function.

    To account for both Pareto efficiency and aggregate performance, fitness of chromosome $Q$ is computed as a product involving $rank(Q)$ and the $L_2$ norm of objectives,
    \begin{equation}\mathscr{E}(Q) = \sqrt{\footnotesize{g_1(\small{Q})^{2} + \footnotesize{g_2(\small{Q})^{2} + \footnotesize{g_3(\small{Q})^{2}}}}}.\end{equation} Note that $\mathscr{E}(\cdot) > 0$ since $g_3(Q)$ is always positive. Let $l_\mathscr{E}$ and $u_\mathscr{E}$ denote the minimum and maximum values of $\mathscr{E}(\cdot)$ in generation $G_h$, respectively. For some $\epsilon > 0$, we set $\gamma=\frac{u_\mathscr{E}}{l_{\mathscr{E}}} + \epsilon.$ Fitness is then defined for $Q \in G_h$ as:
     \begin{equation}\label{formula:fitness}
         f(Q) = \gamma^{rank(Q)}\cdot \mathscr{E}(Q)
     \end{equation}
     In this manuscript's experiments, we use $\epsilon=0.01$, but this value could be increased to encourage greater separation between non-domination levels (See Property 2).

     The global \textit{scalarized} optimization problem posed by this fitness function is:
    \begin{equation} \label{scalarization}
        \argmax_{Q \in \mathcal{S}_{(n,k)}} {f}(Q)
    \end{equation}    
    Since $f$ is increasing with respect to rank and $g_1,g_2,g_3$, a solution to (\ref{scalarization}) is likewise a solution to (\ref{problem}).
    \subsection{Algorithmic Properties} \label{sec:algorithm}
    \texttt{EMSCO} begins by generating a random initial population of $|G|$ chromosomes by calling the function $init()$. This function creates solutions by applying the mutation operator to the default one-stage solution ($[0,0,\ldots,0]$). As a result, the beginning population consists of only two-stage solutions.
    
    After the first population has been generated, the main loop begins. At any point, if the number of unique non-dominated solutions equals the initially specified population size $|G|$, population size is incremented by $inc \in \mathbb{Z}_{\geq 0}$.  By default, $inc = 0$, and this setting was used in all experiments. However, problems with very large non-dominated fronts may warrant setting $inc > 0$. To mitigate computational expense incurred by large populations, if an increased population size is no longer necessary in later generations, $|G|$ is decremented (lines 13-14 in  \texttt{EMSCO} pseudocode).
    
    Three halting conditions are checked before each iteration by calling the function, $converged()$, which returns true if any of the following conditions are satisfied---(i) the maximum number of iterations ($max\_iter$) have been executed, (ii) the highest-scoring chromosome has remained constant for the past $g$ generations, or (iii) $inc = 0$ and the number of unique non-dominated solutions  ($elite\_size$) is equal to population size $|G|$. If none of these halting conditions are true, the loop proceeds, and the next population of solutions is produced using elitism, selection, recombination, and mutation as described in Section \ref{sec:operators}. When the loop ends, the list of all non-dominated solutions in the final generation is returned, sorted by fitness.  We use $\mathscr{N}(G_h)$ to denote the set of all non-dominated solutions in $G_h$.
    \begin{algorithm}
    \begin{footnotesize}
    \SetAlgoLined
    $G_0 \gets init()$;\\ 
    $|G|_{cpy} \gets |G|;$\\
    $h \gets 0;$\\
    \While{$\mathbf{not}$ $converged()$} {
     $rank(G_h)$;\\
     $sort\left(G_h,key=f(\cdot)\right)$;\\
     $elite\_size \gets \max\left(round\left( b\cdot|G_h^*|\right), |E_0^*|\right)$;\\
     $elite\_pop \gets G_h^*\left[0:elite\_size\right]$;\\
     $G_{h+1} \gets_{append} elite\_pop$;\\
     \If{$elite\_size = |G|$}
     {$|G| \pluseq inc;$}
          \If{$elite\_size < (|G| - inc)~ \mathbf{and}~ |G| > |G|_{cpy}$}
     {$|G| = |G| - inc;$}
     \While{$|G_{h+1}| < |G|$} 
     {
     $P_A,P_B \gets select()$;\\
     $\Large{[}C\Large{]} \gets recombine(\Large{[}P_A\Large{]},\Large{[}P_B\Large{]});$\\
     $\Large{[}C\Large{]} \gets mutate(\Large{[}C\Large{]})$;\\
     \If{$gaps(\Large{[}C\Large{]})$} {
     $compress(\Large{[}C\Large{]})$;
     }
     $G_h \gets_{append} C$;\\
     }
     $h \gets h+1$;\\}
     \Return{$\mathscr{N}(G_h)$};
    \end{footnotesize}
    \caption{\texttt{EMSCO}}
    \label{alg:emsco}
    \end{algorithm}
    ~\\Several notable properties follow immediately as consequences of the design described heretofore.
    \begin{property}\label{thm:pres}
    \texttt{EMSCO} preserves globally non-dominated solutions and returns them at the terminal generation.
    \end{property}
    The above property follows immediately from the elitism protocol described in Section \ref{sec:elitism}), since any globally non-dominated solution is likewise non-dominated in any subset of the population. We can then conclude that if any globally non-dominated solution $D \in \mathscr{N}(\mathcal{S}_{(n,k)})$ is encountered in any generation, it is guaranteed to be returned at the terminal generation.
    
    Another notable property of \texttt{EMSCO} regards its incorporation of both non-domination level and scalarized performance, $\mathscr{E}(\cdot)$, into the fitness function that determines probability of selection.
    \begin{property}
     Let $Q \in G_h$, and let probability of selection be denoted as $\mathbb{P}_{_{\tiny{sel.}}}\Large{(}Q\Large{)}$. For chromosomes $R_A, R_B \in G_h$: 

     $$rank(R_A) < rank(R_B) \implies \mathbb{P}_{_{\tiny{sel.}}}(R_A) < \mathbb{P}_{_{\tiny{sel.}}}(R_B),$$
    even if  $\mathscr{E}(R_A) > \mathscr{E}(R_B)$. (ii) Within a particular rank, probability of selection is strictly increasing with respect to $\mathscr{E}(Q)$.
    \end{property}
    \begin{proof}
     (i) This property is a consequence of fitness proportionate selection and the construction of $f(\cdot)$ detailed in Section \ref{sec:fitness}. Let $R_A,R_B$ be chromosomes in generation $G_h$ such that $r_B = rank(R_B) > r_A = rank(R_A).$ Because $\frac{u_\mathscr{E}}{l_{\mathscr{E}}}$ maximizes the ratio between any two $\mathscr{E}(R_A)$ and $\mathscr{E}(R_B)$, we have $$\left(\frac{u_\mathscr{E}}{l_{\mathscr{E}}} + \epsilon\right)^{r_B - r_A} > \frac{\mathscr{E}(R_A)}{\mathscr{E}(R_B)}.$$ Manipulating the above inequality, we obtain
          \begin{equation} \label{fitcrit}\left(\frac{u_\mathscr{E}}{l_{\mathscr{E}}} + \epsilon \right)^{r_A}\mathscr{E}(R_A) < \left(\frac{u_\mathscr{E}}{l_{\mathscr{E}}} + \epsilon \right)^{r_B}\mathscr{E}(R_B).
          \end{equation} By (\ref{formula:fitness}), we then have $f(R_A) < f(R_B).$
     The result (i) then follows from the use of roulette wheel selection in which  $\mathbb{P}_{_{\tiny{sel.}}}\Large{(}Q\Large{)} = \frac{f(Q)}{\sum_{U\in G_h} f(U)}.$ \\(ii).  If $R_A,R_B$ are in the same rank/non-domination level, the $\left(\frac{u_\mathscr{E}}{l_{\mathscr{E}}} + \epsilon\right)^r$ terms cancel in (\ref{fitcrit}). Fitness comparisons between $R_A,R_B$ are then dependent on $\mathscr{E}(\cdot)$ exclusively. (ii) then follows from fitness proportionate selection.
    \end{proof}
    
    It should also be mentioned that separation in fitness (and therefore probability of selection) between each rank decreases as rank decreases. For instance, there is greater separation between fitness scores in the Pareto front and second non-domination level than between fitness scores in the second-to-last and last non-domination levels.

\section{Experiments} \label{sec:exp}
To evaluate \texttt{EMSCO}'s capabilities, experiments are conducted on three data sets from the UCI Machine Learning Repository \cite{dua17} and two synthetic data sets. A variety of confidence thresholds and feature cost schemes are considered.
Experiments in Section \ref{sec:global} empirically evaluate \texttt{EMSCO}'s capacity for global optimization, and experiments in Section \ref{comparisons} compare \texttt{EMSCO} with various budgeted and/or selective classification protocols.

We randomly split the data sets into $50$-$25$-$25$ training, validation, and testing sets. Doing so allows for efficient computation of unbiased estimates for out-of-sample error. 
\texttt{EMSCO} uses the training sets to learn individual classifiers $\{\mathscr{C}(Q_1),\ldots,\mathscr{C}(Q_j),\ldots,\mathscr{C}(Q_{|Q|})\}$.  Validation sets are used to measure $(g_1(Q),g_2(Q),g^*_3(Q))$ during optimization/tuning. The test sets are then used to estimate out-of-sample performance in the comparative experiments.

A simple ``sweep'' \cite{tuning} procedure over a discretized set of selections for $|G|,\hat{m}, \hat{r},\beta, b$ is used to tune \texttt{EMSCO}. A scalarization similar to $\mathscr{E}(\cdot)$ but with a population-independent measure of inverse cost ($(1 - \frac{g_3^*}{\sum \mathcal{C}_i})$) is used to compare performance of parameter combinations. To ensure this manuscript is self-contained, and to promote reproducibility, the parameters selected for each experiment are listed explicitly in the next section. Readers may note that the parameters returned for each instance are quite similar, demonstrating \texttt{EMSCO}'s low variance with regard to parameter changes. In light of this, if tuning is computationally expensive, we suggest $\hat{m}=\frac{1}{n},~ \hat{r}=0.8,~ |G| = 300,~ b=0.2,$ and $\beta=2$ as default parameters.

\subsection{Data Sets}\label{datasets}
Data sets are chosen to provide a robust evaluation of the methods and to include pathologies occurring frequently in real data. Many deployment scenarios in which budgeted learning is applied rely on a fairly small number of sensors/features for classification \cite{trapeznikov2013}, and we accordingly restrict our experiments to data sets with at most fifty features.

 Feature costs are assigned with scaling functions dependent on an assigned \textit{cost class}---an integer value representing the cost incurred while acquiring the respective feature. The cost assignment schemes for each experiment are designed to represent a wide variety of potential deployment scenarios.
Likewise, four distinct confidence thresholds ($\hat{p} = 0.55,0.65,0.75,0.85$) are considered. These values were selected to provide a range of confidence thresholds that represent administrators who are only \textit{mildly} risk averse (low confidence thresholds) to those who are \textit{very} risk averse (high confidence thresholds) and are willing to sacrifice substantial coverage in order to achieve high accuracy.
\subsubsection{Synthetic50 Data Set} \label{synthetic30}
The Synthetic50 data set was constructed using \texttt{sklearn}'s \texttt{make\_classification()} function \cite{pedregosa2011}. The data set consists of 50 features and 4000 records. Binary labels are assigned to the records proportionally (50-50). In this experiment, we account for cases where all features are of roughly equal cost. More precisely, we set all $T_i = 1$, and use scaling function $h(T_i) = 10T_i$. We set \texttt{n\_informative=25}---a parameter of \texttt{make\_classification()} specifying the number of ``informative'', non-redundant features\footnote{For more information regarding sklearn's make\_classification() tool, please refer to \scriptsize{\url{https://scikit-learn.org/stable/modules/generated/sklearn.datasets.make_classification.html}}}. All features are continuous, and four clusters are present in the data (two for each label). In this data set, we account for less cautious settings and set $\hat{p} = 0.55$. Tuning led to mutation rate 0.05, crossover rate 0.80, elite population percent 0.2, population size 300, mutation bias parameter 2.5.
\subsubsection{Synthetic15 Data Set}
The Synthetic15 data set consists of 8000 records and 15 features. 12/15 features are deemed informative using the \texttt{n\_informative} parameter of sklearn's \texttt{make\_classification()} function. Addressing scenarios with a \textit{highly} conservative decision maker willing to sacrifice coverage for improved accuracy, we set $\hat{p} = 0.85$. To account for another possible cost dynamic, we do not assign cost \textit{classes} for this data set, but instead use a noisy, rounded function of the Gini importance (denoted in the following equation as $z_i$) for $\mathcal{F}_i$. Costs for each feature are assigned as $\mathcal{C}_i = \max(1,round\left(100z_i + randint(-1,1)\right)),$ where $randint(-1,1)$ returns either $1$ or $-1$ with equal probability. This procedure yields cost set $\mathcal{C} = [6,8,4,8,8,1,9,10,6,1,9,9,4,1,7]$. 

 As done for the Synthetic50 data set, we use a balanced class distribution ($50\%$ of records have label `1', and $50\%$ have label `0') and two clusters for \textit{each} label. Furthermore, to make the classification task more challenging, we increase proximity between the two possible classes by setting $\texttt{class\_sep} = 0.85$ and assign random labels to $2\%$ of the data.  Mutation rate 0.075, crossover rate 0.80, elite population percent 0.2, population size 250, and mutation bias parameter 2.0 were returned by the tuning procedure.

\begin{figure}
    \centering
    \includegraphics[scale=0.33]{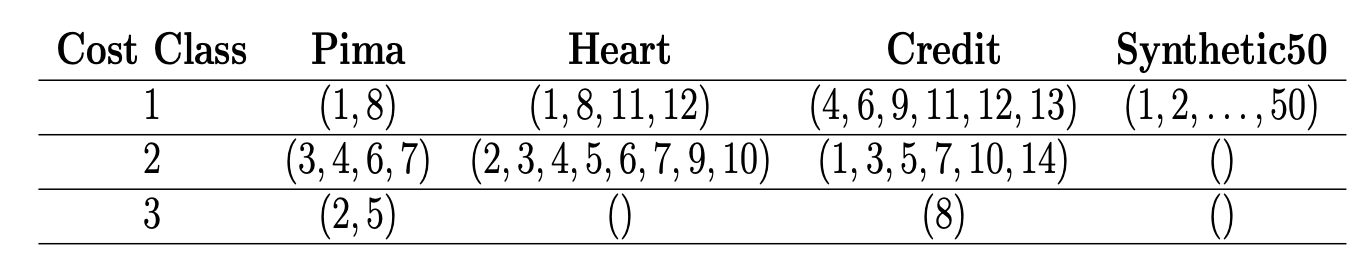}
    \caption*{\footnotesize{\textit{Feature Acquisition Cost Classes}. Each entry corresponds to the set of features assigned to the cost class in the leftmost column. For Pima, Credit, and Heart, features are as indexed on the UCI Machine Learning Repository \cite{dua17}. Synthetic15 is excluded from this table since it employs a different protocol for assigning feature costs.}\\}
    \label{fig:my_label}
\end{figure}
\subsubsection{Pima Diabetes Data Set}
The Pima Diabetes data set contains 768 records and is a popular benchmark for evaluating performance of classifiers. Features are binary, integer, or real-valued. Cost classes $T_i \in \{1,2,3\}$ were assigned to each feature depending on the time and complications inherent in conducting the individual tests. For example, `Age' is assigned $T_i = 1$, and `Plasma glucose concentration' is assigned $T_i = 3$. For this data set, we use the linear scaling function $h(T_i) = 100T_i$. The tuning procedure returned mutation rate 0.075, crossover rate 0.80, elite population percent 0.2, population size 300, mutation bias parameter 2. A 0.65 confidence threshold is used for this data set to account for moderately strict classification instances. We refer to this data set as ``Pima'' for the remainder of this paper.
\subsubsection{Australian Credit Approval (Statlog) Data Set}
This data set is available on the UCI Machine Learning Repository \cite{dua17} and consists of 14 features given in 690 credit applications to a large bank. The target for this data set is binary. Feature descriptions are not provided given the sensitive nature of the data, but the values are designated as continuous or categorical. This data set contains missing values.

Three cost classes are assigned to the features according to their Gini importance \cite{Louppe}, where features of greater importance are assigned greater cost classes. To account for scenarios with greatly varying feature costs, the cost-scaling function is set as $h(T_i) = 10^{T_i}$. Mutation rate 0.075, crossover rate 0.80, elite population percent 0.2, population size 300, and mutation bias parameter 2.5 were returned by the sweep tuning procedure. A 0.75 confidence threshold is applied to predictions. This data set is referred to as ``Credit'' in this paper.

\subsubsection{Heart Failure Clinical Records}
This data set is available on the UCI Machine Learning Repository and was compiled from 300 patients~\cite{dua17}. The data set consists of~$12$ pertinent clinical features that are a mix of categorical, integer, and real values. With the guidance of the data set's author, two time classes were assigned to each feature representing the acquisition cost. For this data set, the cost-scaling function is set as $h(T_i) = 10^{T_i}$, where $T_i$ denotes the cost class ('1' or '2') of the $i$th feature. Mutation rate ($\hat{m}$) 0.075, crossover rate ($\hat{r}$) 0.75, elite population percent ($b$) 0.2, population size ($|G|$) 300, and mutation bias parameter ($\beta$) 2.0 were returned by the tuning procedure.  To account for conservative classification environments, a 0.75 confidence threshold is used to determine early-exit and reject decisions. We refer to this data set as ``Heart'' for the duration of this paper.

\begin{figure*}
\label{fig:globalexperiment}
\begin{center}
\resizebox{\textwidth}{!}{%
\begin{tabular}{ccc}
{\includegraphics[scale=0.9]{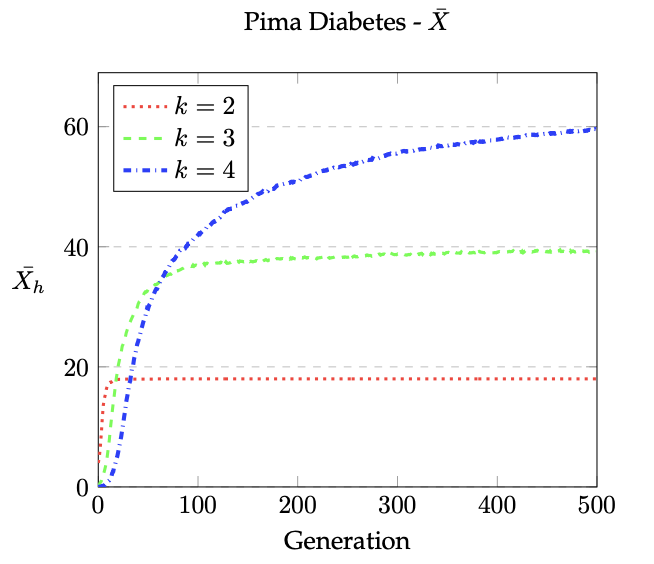}} & 
{\includegraphics[scale=0.9]{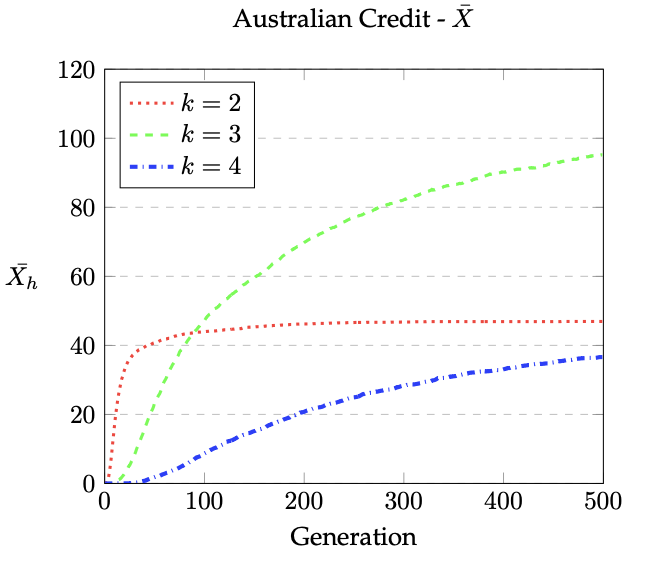}}&
{\includegraphics[scale=0.9]{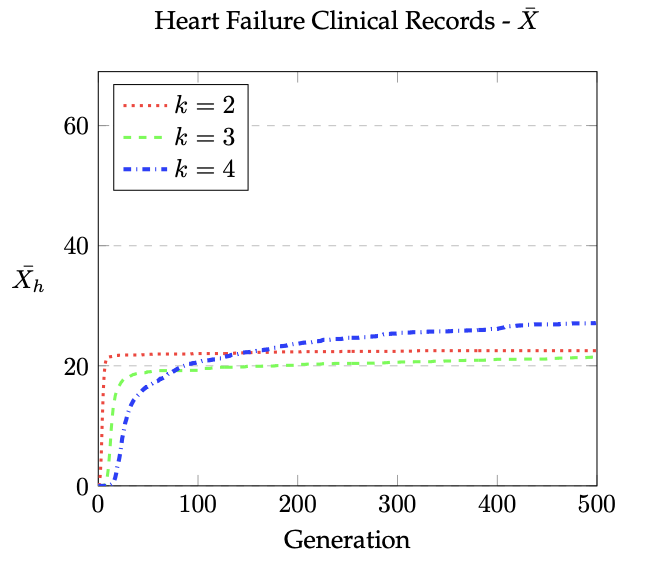}}
\end{tabular}
}
\caption{\small{\textit{Unique, Globally Non-Dominated Solutions at each $h$\textsuperscript{th} Generation.} $\Bar{X_h}$ denotes the sample mean (50 trials) for the count of unique globally optimal solutions present in \texttt{EMSCO}'s $h$th generation during optimization.}}
\label{fig:global}
\end{center}

\end{figure*}
\begin{table}
\begin{center}
\renewcommand{\arraystretch}{1.0}
\begin{tabular}{cccc}
\textbf{Data Set} & \shortstack{\textbf{k=2}} & \shortstack{\textbf{k=3}} &  \shortstack{\textbf{k=4}} \\  \hline
Heart & 24 & 28 & 33\\ \hline
Credit & 47 & 141 & 171\\ \hline
Pima & 18 & 42 & 65\\ \hline
\end{tabular}
\label{tab:import2}  
\caption{Global Pareto Frontier Sizes\label{tab:frontiers}}
\end{center}
\end{table}
\subsection{Global Optimization Experiment} \label{sec:global}
    These experiments assess \texttt{EMSCO}'s ability to find global optima in a large solution space during the learning phase. This first requires establishing a ``ground truth'' fitness landscape from which we can derive the Pareto frontier to use as a point of reference for the solutions in \texttt{EMSCO}'s populations. To this end, we measure $(g_1(Q),g_2(Q),g_3(Q))$ for all  $Q \in \mathcal{S}_{(n,k)}$. Since this represents a significant computational burden, we consider only the Heart, Credit, and Pima data sets with $k \leq 4$. Running these experiments took roughly three weeks with a 32-core Intel Xeon E5-2650 and 256G RAM.

    After establishing the fitness landscape and computing the respective Pareto front for each data set and stage-count ($k=2\ldots4)$, \texttt{EMSCO} was run 50 times. In every run, at each $h$\textsuperscript{th} generation, the number of unique globally optimal solutions $X_h$ (determined by comparison with the predetermined Pareto front) was recorded. These values were then averaged with respect to the 50 samples to obtain $\Bar{X}_h$ and depicted in Figure \ref{fig:global}.

     \texttt{EMSCO} shows itself capable of quickly finding globally Pareto optimal solutions in very large search spaces (e.g., Credit $k=4$ with 250 million solutions). In the Heart and Pima experiments, \texttt{EMSCO} attained nearly all global optima---an encouraging result when considering the Heart, $k=4$ search space contains nearly 15 million total solutions and only 33 globally non-dominated solutions (Table \ref{tab:frontiers}). Another notable pattern evident in these results is the monotonicity of the curves, illustrating Property I empirically.

     \subsection{Comparative Experiment Design} \label{comparisons}
     These experiments evaluate the performance of $\texttt{EMSCO}$ against related approaches. To the best of our knowledge, \texttt{EMSCO} is the first \textit{budgeted} classification method that leverages prediction confidence to improve accuracy by rejecting uncertain predictions. A direct comparison between \texttt{EMSCO} and well-established alternative methods of the same exact nature is therefore not possible. To argue the merit of our perspective and \texttt{EMSCO} specifically, we then compare against a variety of closely-related methods.
     
     \textit{Greedy Miser} (GM) \cite{xu2012} is a variant of stage-wise regression \cite{friedman2001} with a feature-budgeted loss function. In this method, regression trees are added iteratively to form a cost-effective ensemble classifier. \textit{Cost-Sensitive Tree of Classifiers} (CSTC) \cite{xu2014} builds a tree of classifiers optimized for a specific sub-partition of the input space. The aim is to ensure that inputs are classified using only the most pertinent features defined for particular regions of the input space. Doing so reduces unnecessary feature extraction while maintaining accuracy. GM and CSTC do not apply reject options as these methods are non-selective by design and their confidence scores do not represent true class probabilities\footnote{In contrast, \texttt{EMSCO} utilizes logistic regression at each stage which is considered to be a well-calibrated model with interpretable class probabilities \cite{Platt}}.
     As done in their respective papers, GM and CSTC's parameters were tuned during validation using grid search on a large discretized set.

     A \textit{Cost-Ordered $T$-Stage Classifier} (CO-T) is considered as a heuristic utilizing a popular cost-ordered perspective for ordering features in budgeted systems \cite{trapeznikov2013,wang2014}. In this setup, a stage is added for each of the $T$ increasing cost classes. A confidence-based reject option at the final stage is applied. 
     
     We also consider  Cost-and-Coverage-Tuned LASSO method (CaCT LASSO) that performs feature selection during training and then evaluates all inputs in a single stage at test-time with the chosen features before applying a confidence-based reject decision. Cost is then computed as the sum of $C_i$ such that the respective regression coefficient for feature $\mathcal{F}_i$ is non-zero. This method is tuned on the validation set for $\lambda \in \{0, 0.1, 0.2, 0.3,\ldots,10\}$ while accounting for accuracy, cost, and coverage. Note, CO-T and CaCT LASSO methods apply logistic regression at each stage to classify inputs.

     \subsubsection{\texttt{EMSCO} Performance Measurement}
       Experiments for each data set consist of 50 EA runs. For each of these runs, at the terminal generation, the chromosome with maximum fitness is returned and appended to a list. When the 50 EA runs are finished, the chromosomes comprising this list are used to compute average $g_1,g_2,g^*_3$ values \textit{on the test set}, marking the performance listed in Table \ref{Tab:Comp}. Due to the stochastic nature of evolutionary algorithms, a $95\%$ margin of error ($\epsilon_{EA}$) for performance is displayed in Table III. Note, \texttt{EMSCO} is run with the parameters listed in Section \ref{datasets} and $max\_iter=150, k=\min (round(\frac{n}{2}),10)$.
       Average feature acquisition cost ($g_3^*$ as defined in Section \ref{sec:cost}) over the test set is used to measure efficiency of the methods, since $g_3$ only carries meaning within a particular population of solutions.

    \subsubsection{Comparison Experiment Results} \label{sec:compresults}
\begin{table*}
\resizebox{.95\linewidth}{!}{%
\begin{tabular}{ c|ccccc|ccccc|ccccc|} 
 \hline\
  &  & & \textbf{Accuracy\%} & & & & &\textbf{Coverage\%} & & & & &\textbf{Cost}\\
Dataset & EMSCO & GM \cite{xu2012} & CSTC \cite{xu2014} & CaCT Lasso & CO-T  & EMSCO & GM \cite{xu2012} & CSTC \cite{xu2014} & CaCT Lasso & CO-T & EMSCO & GM \cite{xu2012} & CSTC \cite{xu2014}& CaCT Lasso & CO-T\\ [0.5ex] 
 \hline

Pima & $71.1$ & $68.1$ & $67.3$ & $82$ & $72$ & $99$ & $100$ & $100$ & $83.4$ & $97$ & $242.4$ & $500$ & $624$ & $1100$ & $524$ \\ 
 Credit & $83.1$ & $76.4$ & $78.1$ & $89.1$ & $81.3$ & $96.1$ & $100$ & $100$ & $80.1$ & $93.2$ & $365.89$ & $107$ & $522$ & $1420$ & $695.8$ \\
 Heart & $83.4$ & $79.2$ & $81.1$ & $88$ & $87.1$ & $89.3$ & $100$ & $100$ & $65$ & $72$ & $102.4$ & $38.6$ & $560$ & $120$ & $255.3$ \\
Synthetic50 & $66.4$ & $65$ & $64.3$ & $82.6$ & $77.6$ & $100$ & $100$ & $100$ & $92.7$ & $97.5$ & $16.47$ & $76.5$ & $72.2$ & $70$ & $500$ \\ 
Synthetic15 & $90$ & $81.2$ & $85.3$ & $92.4$ & $91.2$ & $66$ & $100$ & $100$ & $52.3$ & $58.5$ & $56.57$ & $75.7$ & $61.3$& $60$ & $79.7$ \\ \hline

\end{tabular}}
\caption{\textit{Comparative Experiment Results}}
\label{Tab:Comp}
\end{table*}
\begin{table}[h]
\begin{center}
\begin{footnotesize}
\renewcommand{\arraystretch}{1.0}
\begin{tabular}{cccc}
\textbf{Data Set} & \textbf{Accuracy} & \textbf{Coverage} &  \textbf{Cost} \\  \hline
Pima & $0.5\%$ & $0.3\%$ & 8.2\\ \hline
Credit & $0.7\%$ & $0.1\%$ & 17.7\\ \hline
Heart & $0.4\%$ & $1.2\%$ & 5.1\\ \hline
Synthetic50 & $0.6\%$ & $0.1\%$ & 4.3\\ \hline
Synthetic15 & $0.1\%$ & $0.1\%$ & 2.4\\ \hline
\end{tabular}
\caption{\small{95\% margin of error values for \texttt{EMSCO} in Table \ref{Tab:Comp}.}}
\end{footnotesize}
\end{center}
\label{Tab:Margins}  
\end{table}
Objective values achieved by each method are listed in Table \ref{Tab:Comp}. Among others, a few notable results arise:
\begin{itemize}
    \item \texttt{EMSCO} is non-dominated in all experiments, performing better than alternatives in at least one objective. Moreover, \texttt{EMSCO} Pareto dominates GM and CSTC in the Synthetic50 experiment and nearly dominates these methods in the Pima experiment as well, if not for the $1\%$ reduction in coverage.
    \item \texttt{EMSCO} offers greater accuracy than the budgeted benchmarks (GM and CSTC) in every experiment. Likewise, in 3/5 experiments (Pima, Synthetic15, Synthetic50), \texttt{EMSCO} also induces substantially lower processing cost than these methods.
    \item \texttt{EMSCO} surpasses every method in at least 2/3 objectives in a majority of experiments (Pima, Synthetic15, and Synthetic50).
    
\end{itemize}
Such results suggest that \texttt{EMSCO} is effective in increasing accuracy compared to non-selective budgeted methods while maintaining strong coverage and low cost. However, it is worth noting that in the Credit and Heart experiments, GM offers considerably lower processing cost. In less conservative settings where accuracy is not paramount, this performance could be favored over \texttt{EMSCO}'s depending on the decision maker's preferences.
\section{Conclusion}
We have introduced \texttt{EMSCO} as a novel approach to manage the objectives of both \textit{selective and budgeted} classification. Experiments conducted on a variety of data sets, confidence thresholds, and cost assignment protocols suggest that the proposed method is capable of finding and maintaining global optima in large solution spaces. Additionally, in multiple experiments, \texttt{EMSCO} is able to simultaneously offer lower cost and greater accuracy than popular budgeted benchmarks.

Future work may consider adding a feature selection option to \texttt{EMSCO} that alters the chromosome representation to include a value (e.g., `$-1$') signifying the respective feature should be excluded. Doing so may promote model simplicity and improve scaling to high-dimensional data. Additionally, since real-world scenarios often impose disproportionate penalties for false-positives and false negatives, use of distinct confidence thresholds (one for each possible label) may prove useful. 
\bibliographystyle{IEEEtran}
\bibliography{reference.bib}

\begin{thebibliography}{10}
\providecommand{\url}[1]{#1}
\csname url@samestyle\endcsname
\providecommand{\newblock}{\relax}
\providecommand{\bibinfo}[2]{#2}
\providecommand{\BIBentrySTDinterwordspacing}{\spaceskip=0pt\relax}
\providecommand{\BIBentryALTinterwordstretchfactor}{4}
\providecommand{\BIBentryALTinterwordspacing}{\spaceskip=\fontdimen2\font plus
\BIBentryALTinterwordstretchfactor\fontdimen3\font minus
  \fontdimen4\font\relax}
\providecommand{\BIBforeignlanguage}[2]{{%
\expandafter\ifx\csname l@#1\endcsname\relax
\typeout{** WARNING: IEEEtran.bst: No hyphenation pattern has been}%
\typeout{** loaded for the language `#1'. Using the pattern for}%
\typeout{** the default language instead.}%
\else
\language=\csname l@#1\endcsname
\fi
#2}}
\providecommand{\BIBdecl}{\relax}
\BIBdecl

\bibitem{ji2007}
\BIBentryALTinterwordspacing
S.~Ji and L.~Carin, ``Cost-sensitive feature acquisition and classification,''
  \emph{Pattern Recogn.}, vol.~40, no.~5, p. 1474–1485, May 2007. [Online].
  Available: \url{https://doi.org/10.1016/j.patcog.2006.11.008}
\BIBentrySTDinterwordspacing

\bibitem{trapeznikov2013}
\BIBentryALTinterwordspacing
K.~Trapeznikov, V.~Saligrama, and D.~Casta{\~{n}}{\'{o}}n, ``Multi-stage
  classifier design,'' \emph{Machine Learning}, vol.~92, no. 2-3, pp. 479--502,
  May 2013. [Online]. Available:
  \url{https://doi.org/10.1007/s10994-013-5349-4}
\BIBentrySTDinterwordspacing

\bibitem{xu2012}
Z.~Xu, K.~Weinberger, and O.~Chapelle, ``The greedy miser: Learning under
  test-time budgets,'' 2012.

\bibitem{kusner}
M.~J. Kusner, W.~Chen, Q.~Zhou, Z.~Xu, K.~Q. Weinberger, and Y.~Chen,
  ``Feature-cost sensitive learning with submodular trees of classifiers,'' in
  \emph{Proceedings of the Twenty-Eighth AAAI Conference on Artificial
  Intelligence}, ser. AAAI'14.\hskip 1em plus 0.5em minus 0.4em\relax AAAI
  Press, 2014, p. 1939–1945.

\bibitem{xu2014}
\BIBentryALTinterwordspacing
Z.~E. Xu, M.~J. Kusner, K.~Q. Weinberger, M.~Chen, and O.~Chapelle,
  ``Classifier cascades and trees for minimizing feature evaluation cost,''
  \emph{Journal of Machine Learning Research}, vol.~15, no.~62, pp. 2113--2144,
  2014. [Online]. Available: \url{http://jmlr.org/papers/v15/xu14a.html}
\BIBentrySTDinterwordspacing

\bibitem{nan16}
F.~Nan, J.~Wang, and V.~Saligrama, ``Pruning random forests for prediction on a
  budget,'' in \emph{Advances in Neural Information Processing Systems 29:
  NeurIPS 2016, Barcelona, Spain}, 2016, pp. 2334--2342.

\bibitem{wang2014}
\BIBentryALTinterwordspacing
J.~Wang, K.~Trapeznikov, and V.~Saligrama, ``{An LP for Sequential Learning
  Under Budgets},'' vol.~33.\hskip 1em plus 0.5em minus 0.4em\relax PMLR,
  22--25 Apr 2014, pp. 987--995. [Online]. Available:
  \url{http://proceedings.mlr.press/v33/wang14b.html}
\BIBentrySTDinterwordspacing

\bibitem{hamilton20202}
N.~H. Hamilton, S.~McKinney, E.~Allan, and E.~W. Fulp, ``An efficient
  multi-stage approach for identifying domain shadowing,'' in \emph{ICC 2020 -
  2020 IEEE International Conference on Communications (ICC)}, 2020, pp. 1--7.

\bibitem{nogueira2019}
R.~Nogueira, W.~Yang, K.~Cho, and J.~Lin, ``Multi-stage document ranking with
  bert,'' 2019.

\bibitem{jansich}
\BIBentryALTinterwordspacing
J.~Janisch, T.~Pevn{\'{y}}, and V.~Lis{\'{y}}, ``Classification with costly
  features as a sequential decision-making problem,'' \emph{CoRR}, vol.
  abs/1909.02564, 2019. [Online]. Available:
  \url{http://arxiv.org/abs/1909.02564}
\BIBentrySTDinterwordspacing

\bibitem{viola}
\BIBentryALTinterwordspacing
P.~A. Viola and M.~J. Jones, ``Rapid object detection using a boosted cascade
  of simple features.'' in \emph{CVPR (1)}, 2001. [Online]. Available:
  \url{http://dblp.uni-trier.de/db/conf/cvpr/cvpr2001-1.htmlViolaJ01}
\BIBentrySTDinterwordspacing

\bibitem{friedman2001}
\BIBentryALTinterwordspacing
J.~H. Friedman, ``{Greedy function approximation: A gradient boosting
  machine.}'' \emph{The Annals of Statistics}, vol.~29, no.~5, pp. 1189 --
  1232, 2001. [Online]. Available: \url{https://doi.org/10.1214/aos/1013203451}
\BIBentrySTDinterwordspacing

\bibitem{nan2017}
\BIBentryALTinterwordspacing
F.~Nan and V.~Saligrama, ``Adaptive classification for prediction under a
  budget,'' in \emph{Advances in Neural Information Processing Systems 30},
  I.~Guyon, U.~V. Luxburg, S.~Bengio, H.~Wallach, R.~Fergus, S.~Vishwanathan,
  and R.~Garnett, Eds.\hskip 1em plus 0.5em minus 0.4em\relax Curran
  Associates, Inc., 2017, pp. 4730--4740. [Online]. Available:
  \url{http://papers.nips.cc/paper/7058-adaptive-classification-for-prediction-under-a-budget.pdf}
\BIBentrySTDinterwordspacing

\bibitem{andrade20}
\BIBentryALTinterwordspacing
D.~Andrade and Y.~Okajima, ``Adaptive covariate acquisition for minimizing
  total cost of classification,'' 2020. [Online]. Available:
  \url{https://arxiv.org/abs/2002.09162}
\BIBentrySTDinterwordspacing

\bibitem{yaniv10}
\BIBentryALTinterwordspacing
R.~El-Yaniv and Y.~Wiener, ``On the foundations of noise-free selective
  classification,'' \emph{Journal of Machine Learning Research}, vol.~11,
  no.~53, pp. 1605--1641, 2010. [Online]. Available:
  \url{http://jmlr.org/papers/v11/el-yaniv10a.html}
\BIBentrySTDinterwordspacing

\bibitem{geifman2017}
Y.~Geifman and R.~El-Yaniv, ``Selective classification for deep neural
  networks,'' 2017.

\bibitem{deb01}
K.~Deb, \emph{Multi-Objective Optimization Using Evolutionary
  Algorithms}.\hskip 1em plus 0.5em minus 0.4em\relax New York, NY, USA: John
  Wiley \& Sons, Inc., 2001.

\bibitem{cancerconf}
\BIBentryALTinterwordspacing
D.~Lichtblau and C.~Stoean, ``Cancer diagnosis through a tandem of classifiers
  for digitized histopathological slides,'' \emph{PLoS One}, 2019. [Online].
  Available: \url{https://pubmed.ncbi.nlm.nih.gov/30650087/}
\BIBentrySTDinterwordspacing

\bibitem{patel}
B.~N. Patel, L.~B. Rosenberg, G.~Willcox, D.~Baltaxe, M.~Lyons, J.~A. Irvin,
  P.~Rajpurkar, T.~J. Amrhein, R.~Gupta, S.~S. Halabi, C.~Langlotz, E.~Lo,
  J.~G. Mammarappallil, A.~J. Mariano, G.~Riley, J.~Seekins, L.~Shen,
  E.~Zucker, and M.~P. Lungren, ``Human–machine partnership with artificial
  intelligence for chest radiograph diagnosis,'' \emph{NPJ Digital Medicine},
  vol.~2, 2019.

\bibitem{eiben}
A.~Eiben and J.~Smith, \emph{\BIBforeignlanguage{English}{Introduction to
  Evolutionary Computing}}, ser. Natural Computing Series.\hskip 1em plus 0.5em
  minus 0.4em\relax Springer, 2015, gebeurtenis: 2nd edition.

\bibitem{deb02}
K.~Deb, A.~Pratap, S.~Agarwal, and T.~Meyarivan, ``A fast and elitist
  multiobjective genetic algorithm: Nsga-ii,'' \emph{IEEE Transactions on
  Evolutionary Computation}, vol.~6, no.~2, pp. 182--197, 2002.

\bibitem{dua17}
\BIBentryALTinterwordspacing
D.~Dua and C.~Graff, ``{UCI} machine learning repository,'' 2017. [Online].
  Available: \url{http://archive.ics.uci.edu/ml}
\BIBentrySTDinterwordspacing

\bibitem{tuning}
K.~De~Jong, ``Parameter setting in eas: a 30 year perspective,'' in
  \emph{Parameter setting in evolutionary algorithms}.\hskip 1em plus 0.5em
  minus 0.4em\relax Springer, 2007, pp. 1--18.

\bibitem{pedregosa2011}
F.~Pedregosa, G.~Varoquaux, A.~Gramfort, V.~Michel, B.~Thirion, O.~Grisel,
  M.~Blondel, P.~Prettenhofer, R.~Weiss, V.~Dubourg \emph{et~al.},
  ``Scikit-learn: Machine learning in python,'' \emph{Journal of machine
  learning research}, vol.~12, no. Oct, pp. 2825--2830, 2011.

\bibitem{Louppe}
\BIBentryALTinterwordspacing
G.~Louppe, L.~Wehenkel, A.~Sutera, and P.~Geurts, ``Understanding variable
  importances in forests of randomized trees,'' in \emph{Advances in Neural
  Information Processing Systems}, vol.~26, 2013. [Online]. Available:
  \url{https://proceedings.neurips.cc/paper/2013/file/e3796ae838835da0b6f6ea37bcf8bcb7-Paper.pdf}
\BIBentrySTDinterwordspacing

\bibitem{Platt}
J.~C. Platt, ``Probabilistic outputs for support vector machines and
  comparisons to regularized likelihood methods,'' in \emph{ADVANCES IN LARGE
  MARGIN CLASSIFIERS}.\hskip 1em plus 0.5em minus 0.4em\relax MIT Press, 1999,
  pp. 61--74.

\end{thebibliography}
    
\end{document}